\documentclass{article}

\usepackage[preprint]{neurips_2026}

\usepackage[utf8]{inputenc} 
\usepackage[T1]{fontenc}    
\usepackage{hyperref}       
\usepackage{url}            
\usepackage{booktabs}       
\usepackage{amsfonts}       
\usepackage{nicefrac}       
\usepackage{microtype}      
\usepackage{xcolor}         

\usepackage{graphicx}
\usepackage{subcaption}
\usepackage{amsmath}
\usepackage{amssymb}
\usepackage{mathtools}
\usepackage{amsthm}
\usepackage{algorithm}
\usepackage{algorithmic}
\usepackage{wrapfig}

\title{Importance Weighting for Unlabeled-unlabeled Learning under Distribution Shift}

\author{
\textbf{Atsutoshi Kumagai},
\textbf{Tomoharu Iwata},
\textbf{Hiroshi Takahashi}, \\
\textbf{Taishi Nishiyama}, 
\textbf{Kazuki Adachi},
\textbf{Yasuhiro Fujiwara} \\
NTT, Inc. \\
\texttt{atsutoshi.kumagai@ntt.com}
}

\begin{document}

\maketitle

\begin{abstract}
Unlabeled-unlabeled (UU) learning allows us to learn a binary classifier from two sets of unlabeled data with different class-priors.
It is a general framework because it includes a wide variety of supervised learning such as positive-unlabeled (PU) learning, noisy label learning, and similarity-based learning.
Existing UU learning assumes that the test and training distributions have the same class-conditional densities.
However, this assumption rarely holds in practice due to distribution shifts.
This paper proposes a distribution shift adaptation method for UU learning that uses UU data in the training distribution and a few UU data in the test distribution.
The proposed method is based on the importance weighting, which minimizes the test risk by using training data with estimated importance weights.
Although existing importance weighting methods cannot handle UU data, 
we show that it can be done in a principled manner.
Thanks to the generality of UU learning,
our method can handle various learning problems such as PU and noisy label learning 
under distribution shift within a single framework
while existing methods are usually tailored to a specific problem.
Moreover, it does not require any assumption of the shift types such as covariate shift.
We experimentally demonstrate the effectiveness of the proposed method with real-world datasets.
\end{abstract}

\section{Introduction}

In supervised learning, a large amount of labeled data is required for learning accurate classifiers.
However, in practice, labeled data are often very costly or even infeasible to collect.
This fact has motivated us to investigate learning algorithms that work well from data with weak supervision (e.g., noisy labels), which are often easier to collect than clean and complete labeled data
\cite{sugiyama2022machine}.
In this paper, we consider unlabeled-unlabeled (UU) learning where the aim is to learn a binary classifier from two unlabeled datasets that share the class-conditional densities but have different class-priors (i.e., the proportion of positives in each unlabeled data)
\cite{lu2018minimal,lu2020mitigating,lu2021binary}.
In this setting, class-priors can be regarded as weak supervision.
We focus on this challenging setting because of its high generality: it can represent a wide variety of learning problems, such as positive-negative (PN) learning~\cite{vapnik1998statistical}, 
positive-unlabeled (PU) learning
\cite{du2015convex,kiryo2017positive,bekker2020learning}, 
noisy label learning
\cite{xie2024weakly,charoenphakdee2019symmetric},
pairwise-comparison learning~\cite{feng2021pointwise}, 
and similarity-based learning~\cite{bao2018classification,shimada2021classification}, 
by specifying values of the class-priors~\cite{chiang2025unified}.

Although several UU learning methods have recently been proposed
\cite{lu2018minimal,lu2020mitigating,lu2021binary,wu2023making}, 
they assume that the test and training distributions have the same class-conditional densities.
However, in practice, this assumption is often violated by various factors 
such as differences in data collection periods
\cite{kumagaiimportance}
and environments
\cite{kumagai2019transfer}.
For example, in cyber security, (noisy) labeled malicious and normal data can be regarded as UU data, and the data distribution can vary over time by the emergence of new types of attacks
\cite{kumagai2024auc}.
When such a distribution shift occurs, the performances of existing methods drastically deteriorate.
\begin{figure*}[t]
    \centering
    \includegraphics[width=11.0cm]{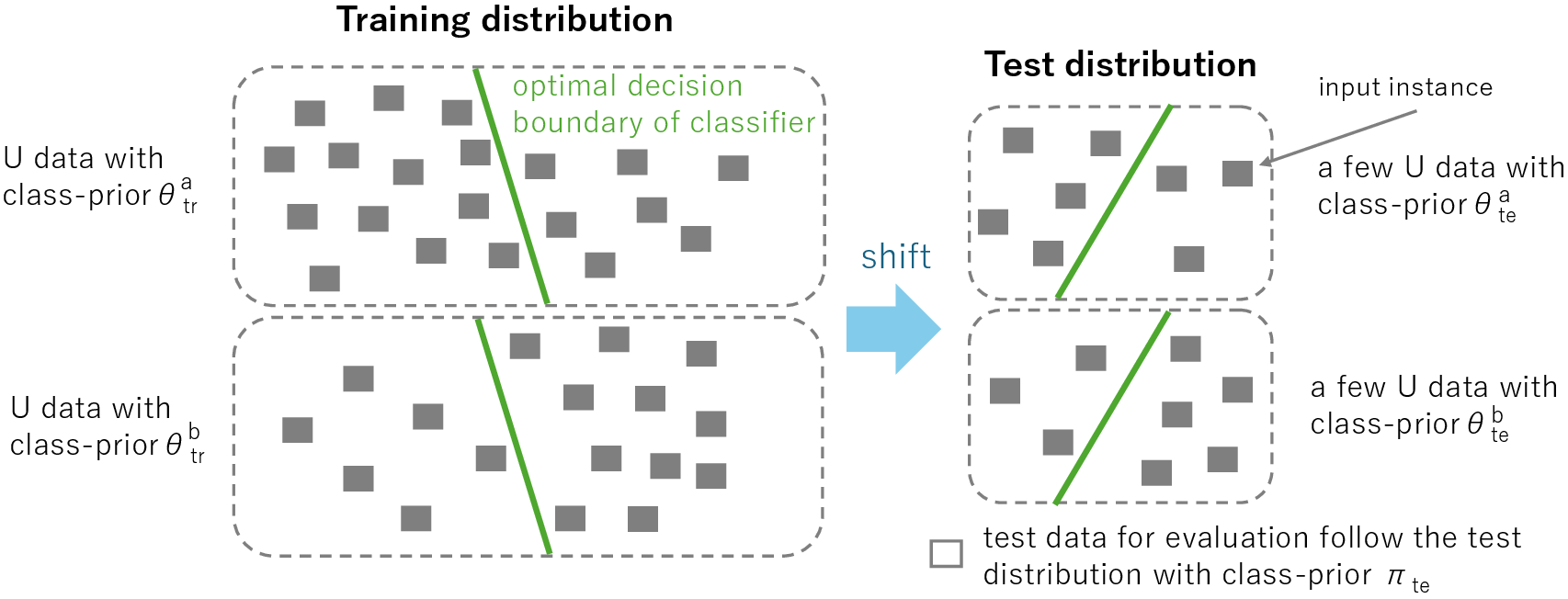} 
    \caption{Our problem setting. We are given UU data in the training distribution and a few UU data in the test distribution.
    UU data in each distribution shares the class-conditional densities but have different class-priors (i.e., $\theta^{{\rm a}}_{{\rm tr}} \neq \theta^{{\rm b}}_{{\rm tr}}$ and $\theta^{{\rm a}}_{{\rm te}} \neq \theta^{{\rm b}}_{{\rm te}}$).
    The class-conditional densities can vary between the training and test distributions.
    The aim is to learn a binary classifier that can accurately classify test data drawn from the test distribution with true class-prior $\pi_{{\rm te}}$ by using these UU datasets. More details of our problem setting are provided in subsection \ref{probset}.}
    \label{fig:overview}
\end{figure*}

In this paper, we propose a distribution shift adaptation method for UU learning that uses UU data in the training distribution and a few UU data in the test distribution\footnote{Although the test (training) distributions are sometimes referred to as the target (source) distributions, this paper uses the former terminology following previous distribution shift adaptation studies~\cite{fang2020rethinking,kumagaiimportance}.}.
A few UU data are relatively easy to collect even in the test distribution.
Figure \ref{fig:overview} shows the overview of our problem setting.
Our method is based on the importance weighting, which is a well-established and commonly used framework for distribution shift adaptation on ordinary supervised learning
\cite{sugiyama2007covariate,fang2024generalizing,lu2022rethinking}.
This framework estimates importance weights, i.e., the ratio between training and test densities, and learns classifiers by minimizing the importance-weighted empirical training risk that is approximately equivalent to the test risk.
Although many distribution shift adaptation methods have been proposed for settings such as PN learning
\cite{fang2020rethinking,fang2024generalizing} and PU learning
\cite{kumagaiimportance},
no existing methods can treat UU data.
To the best of our knowledge, we are the first to show that importance weighting can be performed from UU data in a principled manner.
By minimizing 
a weighted sum of the importance-weighted empirical risk with UU data in the training distribution and the empirical risk with UU data in the test distribution,
the proposed method can train classifiers that fit on the test distribution.

Thanks to the generality of UU learning, the proposed method can handle various types of learning problems under distribution shift within a single framework, including PN, PU, noisy label, pairwise-comparison, and similarity-based learning.
This contrasts with existing distribution shift adaptation methods, which are typically tailored to a specific problem
\cite{pan2009survey,singhal2023domain}.
Note that there are no existing distribution shift adaptation methods that assume noisy labels in both distributions nor methods for similarity-based learning or paired-comparison learning.
In addition, the proposed method can flexibly handle different types of supervision between the training and test phases, e.g., noisy labeled data in the training distribution and a few PN data in the test distribution.
Furthermore, by using UU data in the test distribution, the proposed method does not require assumptions about shift types such as covariate shift
\cite{shimodaira2000improving}
and concept shift
\cite{moreno2012unifying}, 
which are usually required in existing distribution shift adaptation methods that use unlabeled data in the test distribution
\cite{shimodaira2000improving}.
Since shift types are generally difficult or impossible to 
identify from unlabeled data
\cite{kumagaiimportance}, this property is preferable in practice.

\section{Related Work}

Discriminative clustering or unsupervised classification methods attempt to learn a classifier from a set of unlabeled data
\cite{xu2004maximum,krause2010discriminative}.
However, they are often suboptimal since they rely on a clustering assumption that one cluster corresponds to one class, which is often violated in practice
\cite{lu2018minimal,lu2020mitigating}.
To overcome this issue, UU learning methods are based on the empirical risk minimization (ERM) that rewrites the risk using two unlabeled datasets with different class-priors
\cite{lu2018minimal,lu2020mitigating,menon2015learning}.
It can learn a binary classifier without the clustering assumption
\cite{lu2018minimal}.
Some methods can use $m$ $(m \geq 2)$ sets of unlabeled data with different class-priors to learn a binary classifier
\cite{lu2021binary,wu2023making}.
However, all these methods assume that the test and training distributions have the same class-conditional densities, and thus they are inappropriate for our problem settings where the class-conditional densities can vary.

Distribution shift or domain adaptation methods attempt to mitigate the gap between the training and test distributions
\cite{pan2009survey,wilson2020survey,tan2018survey,singhal2023domain}.
Although many methods have been proposed,
there are no methods for UU learning.
For the adaptation,
one representative approach is to learn invariant features between the training and test distributions
\cite{saito2018maximum,motiian2017unified,ganin2015unsupervised,kumagai2019unsupervised,Lee_2019_CVPR}.
Although this approach is promising in some applications, 
the invariant features often do not improve the performance since they do not explicitly minimize the test risk
\citep{zhao2019learning,kumagaiimportance}.
Another representative approach is the importance weighting, which can adapt to the shift by explicitly minimizing the test risk, and it has solid theoretical properties such as consistency 
\citep{lu2022rethinking,sugiyama2012density}.
Accordingly, we extend it to the UU setting.

As for data used for the adaptation, existing methods often assume unlabeled data in the test distribution 
\cite{saito2018maximum,ganin2015unsupervised,kumagai2019unsupervised,shimodaira2000improving}.
Since there is no supervision in the test distribution, they typically assume a specific shift type such as covariate shift
\cite{shimodaira2000improving}.
However, the shift type is generally difficult or impossible 
to identify from unlabeled data
\cite{fang2020rethinking,kumagaiimportance}.
Some recent importance weighting methods can adapt to the shift without shift type assumptions 
by using some supervision in the test distribution, such as a few PN data
\cite{fang2020rethinking,fang2024generalizing}
and a few PU data
\cite{kumagaiimportance}.
The proposed method also does not require shift type assumptions because it uses a few UU data in the test distribution.
By appropriately setting the class-priors, the proposed method can recover these methods
\cite{fang2020rethinking,kumagaiimportance} 
as special cases.
In general, since the proposed method must perform importance weighting solely from unlabeled data, 
it addresses a more challenging problem than these methods that can exploit labeled (positive) data.
Although several methods assume noisy labeled data in the training distribution and unlabeled data in the test distribution
\cite{shu2019transferable,yu2020label,yu2022robust,han2022towards},
they cannot treat noisy labeled data in both the training and test distributions, but the proposed method can.

\section{Preliminary}
\label{prelim}

We explain UU learning based on the ERM
\cite{lu2018minimal,lu2020mitigating}.
Let ${\bf x} \in {\cal X} $ and $y \in \{ \pm{1} \}$ be the input and output random variables, 
where ${\cal X}$ is the input space, $y=+1$ and $-1$ represent positive and negative classes, respectively.
Let $p({\bf x},y)$ be the joint density, $p ^{{\rm p}}({\bf x}):=p({\bf x} |y=+1)$ and $p ^{{\rm n}}({\bf x} ):=p({\bf x} |y=-1)$
be the positive- and negative-conditional densities,
$p({\bf x} ) = \pi p^{{\rm p}}({\bf x}) + (1 - \pi) p^{{\rm n}}({\bf x} )$ be the marginal density, and $\pi :=p(y=+1)$ be the class-prior.
$f: {\cal X} \to \mathbb{R} $ is a decision function and the predicted label is obtained by $t = {\rm sign} (f({\bf x}))$, where ${\rm sign} (\cdot)$ is a sign function.
Let $\ell: \mathbb{R} \times \{ \pm{1} \} \to \mathbb{R} _{\geq 0} $ be the loss function, such that value $\ell(t,y)$ means the loss for ground truth label $y$ by $t$.
The binary classification aims to learn $f$ that minimizes the expected test error, called the risk,
\begin{align}
\label{risk}
R (f) := \mathbb{E} _{p({\bf x},y)} [\ell (f({\bf x}),y)] = \pi \mathbb{E} _{p ^{{\rm p}} ({\bf x}) } [\ell (f({\bf x}),+1)] + (1 \!-\! \pi) \mathbb{E} _{ p ^{{\rm n}} ({\bf x}) } [\ell (f({\bf x}),-1)],
\end{align}
where $\mathbb{E} _{p(z)} $ is an expectation over $p(z)$. 
In standard supervised learning, we are given labeled positive and negative data drawn from $p({\bf x},y)$. Thus, we can directly train the classifier by minimizing the empirical risk that is calculated by replacing the expectation in Eq. \eqref{risk} with the sample average.

However, in UU learning, we are given two sets of unlabeled data $X^{{\rm a}}$ and $X^{{\rm b}}$ drawn from the following marginal distributions:
\begin{align}
&X^{{\rm a}} \!=\! \{ {\bf x}^{{\rm a}} _{n} \}_{n=1}^{N^{{\rm a}}} \!\sim\! p^{{\rm a}} ({\bf x}) \!:=\! \theta^{{\rm a}} p^{{\rm p}} ({\bf x}) \!+\! (1 - \theta^{{\rm a}}) p^{{\rm n}}  ({\bf x}), \nonumber\\
&X^{{\rm b}} \!=\! \{ {\bf x}^{{\rm b}} _{n} \}_{n=1}^{N^{{\rm b}}} \!\sim\! p^{{\rm b}} ({\bf x}) \!:=\! \theta^{{\rm b}} p^{{\rm p}} ({\bf x}) \!+\! (1 - \theta^{{\rm b}}) p^{{\rm n}}  ({\bf x}), \label{mar_pre2}
\end{align}
where $\theta^{{\rm a}}$ and $\theta^{{\rm b}}$ are two class-priors such that $\theta^{{\rm a}} \neq \theta^{{\rm b}}$.
Marginal densities $p({\bf x} )$, $p^{{\rm a}} ({\bf x})$, and $p^{{\rm b}} ({\bf x})$ share the class-conditional densities $p^{{\rm p}} ({\bf x})$ and $p^{{\rm n}} ({\bf x})$.
To train the classifier from the two unlabeled datasets, 
UU learning rewrites the risk in Eq. \eqref{risk} with two marginal densities $p^{{\rm a}} ({\bf x})$ and $p^{{\rm b}} ({\bf x})$.
Specifically, by solving Eq. \eqref{mar_pre2} for  $p ^{{\rm p}}({\bf x})$ and $p ^{{\rm n}}({\bf x})$, we can obtain the following equations:
\begin{align}
\pi p^{{\rm p}} ({\bf x}) = a p^{{\rm a}} ({\bf x}) - c p^{{\rm b}}  ({\bf x}), \ \ (1-\pi) p^{{\rm n}} ({\bf x}) = -b p^{{\rm a}} ({\bf x}) + d p^{{\rm b}}  ({\bf x}), \label{posneg_pre}
\end{align}
where $a := \frac{(1-\theta^{{\rm b}})\pi}{\theta^{{\rm a}}-\theta^{{\rm b}}}$, $b := \frac{\theta^{{\rm b}}(1-\pi)}{\theta^{{\rm a}}-\theta^{{\rm b}}}$, $c := \frac{(1-\theta^{{\rm a}})\pi}{\theta^{{\rm a}}-\theta^{{\rm b}}}$, and $d := \frac{\theta^{{\rm a}}(1-\pi)}{\theta^{{\rm a}}-\theta^{{\rm b}}}$.
By using Eq. \eqref{posneg_pre}, we can obtain
\begin{align}
\pi \mathbb{E} _{p ^{{\rm p}} ({\bf x}) } [\ell (f({\bf x}),+1)] &= a \mathbb{E}_{p ^{{\rm a}} ({\bf x}) } [ \ell(f({\bf x}, +1)] -c \mathbb{E}_{p ^{{\rm b}} ({\bf x}) } [ \ell(f({\bf x}, +1)], \nonumber\\
(1-\pi) \mathbb{E} _{p ^{{\rm n}} ({\bf x}) } [\ell (f({\bf x}),-1)] &= -b \mathbb{E}_{p ^{{\rm a}} ({\bf x}) } [ \ell(f({\bf x}, -1)] + d \mathbb{E}_{p ^{{\rm b}} ({\bf x}) } [ \ell(f({\bf x}, -1)]. \label{expt_pre2}
\end{align}
Plugging them into Eq. \eqref{risk}, we obtain
\begin{align}
\label{uu_risk}
R (f) \!=\! a \mathbb{E}_{p ^{{\rm a}} ({\bf x}) } [ \ell(f({\bf x}), +1)] \!-\! c \mathbb{E}_{p ^{{\rm b}} ({\bf x}) } [ \ell(f({\bf x}), +1)] \!-\! b \mathbb{E}_{p ^{{\rm a}} ({\bf x}) } [ \ell(f({\bf x}), -1)] \!+\! d \mathbb{E}_{p ^{{\rm b}} ({\bf x}) } [ \ell(f({\bf x}), -1)].
\end{align}
Since Eq. \eqref{uu_risk} is represented with $p^{{\rm a}} ({\bf x})$ and $p^{{\rm b}} ({\bf x})$, we can approximate it with $X^{{\rm a}}$ and $X^{{\rm b}}$.
We call this rewritten risk {\it the UU risk}.
Although both equations in \eqref{expt_pre2} are non-negative by definition, when highly expressive models such as neural networks are used for $f$, their empirical estimates can take negative values leading to serious overfitting
\cite{lu2020mitigating}.
To alleviate this issue, non-negative correction such as the absolute value correction is commonly used
\cite{lu2020mitigating,lu2021binary}.
Specifically, the empirical estimate of Eq. \eqref{uu_risk} with the absolute value correction is given by
\begin{align}
\label{uu_risk_cor}
{\hat R} (f) \!=\! \left| \frac{a}{N^{{\rm a}}} \!\sum_{n=1}^{N^{{\rm a}}}\! \ell (f({\bf x} _n^{{\rm a}}), \!+1\!) \!-\!  \!\frac{c}{N^{{\rm b}}}\! \sum_{n=1}^{N^{{\rm b}}} \ell (f({\bf x} _n^{{\rm b}}), \!+1\!) \right| \!+\! \left| \frac{d}{N^{{\rm b}}} \!\sum_{n=1}^{N^{{\rm b}}}\! \ell (f({\bf x} _n^{{\rm b}}), \!-1\!) \!-\! \frac{b}{N^{{\rm a}}} \!\sum_{n=1}^{N^{{\rm a}}}\! \ell (f({\bf x} _n^{{\rm a}}), \!-1\!)  \right|,
\end{align}
where $| \cdot |$ is the absolute value function to penalize the risk for being negative. When there are many UU data, we can obtain an accurate classifier $f$ by minimizing Eq. \eqref{uu_risk_cor} w.r.t. parameters of $f$.
UU learning includes a wide variety of learning problems as special cases. 
For example, when $\theta^{{\rm a}}=1$ and $\theta^{{\rm b}}=0$, $X^{{\rm a}}$ and $X^{{\rm b}}$ are positive and negative data, respectively; Eq. \eqref{uu_risk_cor} becomes the empirical risk on ordinary PN learning with $(a, b, c, d)=(\pi, 0, 0, 1-\pi)$
\cite{vapnik1998statistical}.
When $\theta^{{\rm a}}=1$ and $\theta^{{\rm b}}=\pi$, $X^{{\rm a}}$ and $X^{{\rm b}}$ are positive and unlabeled data, respectively; Eq. \eqref{uu_risk_cor} becomes the non-negative empirical risk on PU learning with $(a, b, c, d)=(\pi, \pi, 0, 1)$
\cite{hammoudeh2020learning}.
When $1 > \theta^{{\rm a}} > \theta^{{\rm b}} > 0$, $X^{{\rm a}}$ and $X^{{\rm b}}$ can be regarded as noisy positive and negative data, respectively; Eq. \eqref{uu_risk_cor} becomes the non-negative empirical risk on noisy label learning.

\section{Proposed Method}
\label{proposed}

We first define our problem setting (subsection~\ref{probset}).
Then, we derive the importance-weighted UU risk (subsection~\ref{sec:uurisk}) and the estimation method for importance weights from UU data (subsection~\ref{sec:iwe}).
After describing our classifier loss function with the importance-weighted UU risk (subsection~\ref{sec:cl}), 
we explain our training procedure (subsection~\ref{train_pro}).

\subsection{Problem Setting}
\label{probset}

Let $p_{{\rm tr}} ({\bf x}, y)$ be the joint density of the training distribution, $p ^{{\rm p}}_{{\rm tr}} ({\bf x}):=p_{{\rm tr}} ({\bf x} |y=+1)$ and $p_{{\rm tr}}^{{\rm n}}({\bf x} ):=p_{{\rm tr}} ({\bf x} |y=-1)$
be the positive and negative-conditional training densities, respectively, and $\pi_{{\rm tr}} :=p_{{\rm tr}} (y=+1)$ be the positive training class-prior.
Similarly, let  $p_{{\rm te}} ({\bf x}, y)$ be the joint density of the test distribution, $p ^{{\rm p}}_{{\rm te}} ({\bf x}):=p_{{\rm te}} ({\bf x} |y=+1)$ and $p_{{\rm te}}^{{\rm n}}({\bf x} ):=p_{{\rm te}} ({\bf x} |y=-1)$
be the positive and negative-conditional test densities, respectively, and $\pi_{{\rm te}} :=p_{{\rm te}} (y=+1)$ be the positive test class-prior.
We assume that the training and test distributions are related but different, $p _{{\rm tr}} ({\bf x}, y) \neq p _{{\rm te}} ({\bf x}, y)$, which is the most general shift form.
Note that we do not know the specific shift type such as covariate shift.

Suppose that we are given two sets of unlabeled data $X^{{\rm a}}_{{\rm tr}}$ and $X^{{\rm b}}_{{\rm tr}}$ drawn from different marginal densities $p^{{\rm a}}_{{\rm tr}} ({\bf x})$ and $p^{{\rm b}}_{{\rm tr}} ({\bf x})$, respectively, in the training phase:
\begin{align}
&X^{{\rm a}}_{{\rm tr}} \!=\! \{ {\bf x}^{{\rm a}} _{{\rm tr}, n} \}_{n=1}^{N^{{\rm a}}_{{\rm tr}}} \!\sim\! p^{{\rm a}}_{{\rm tr}} ({\bf x}) \!:=\! \theta^{{\rm a}}_{{\rm tr}} p^{{\rm p}}_{{\rm tr}} ({\bf x}) \!+\! (1 \!-\! \theta^{{\rm a}}_{{\rm tr}}) p^{{\rm n}}_{{\rm tr}}  ({\bf x}),  \nonumber\\
&X^{{\rm b}}_{{\rm tr}} \!=\! \{ {\bf x}^{{\rm b}} _{{\rm tr}, n} \}_{n=1}^{N^{{\rm b}}_{{\rm tr}}} \!\sim\! p^{{\rm b}}_{{\rm tr}} ({\bf x}) \!:=\! \theta^{{\rm b}}_{{\rm tr}} p^{{\rm p}}_{{\rm tr}} ({\bf x}) \!+\! (1 \!-\! \theta^{{\rm b}}_{{\rm tr}}) p^{{\rm n}}_{{\rm tr}}  ({\bf x}), \label{mar_train2}
\end{align}
where $\theta^{{\rm a}}_{{\rm tr}}$ and $\theta^{{\rm b}}_{{\rm tr}}$ are two class-priors such that $\theta^{{\rm a}}_{{\rm tr}} \neq \theta^{{\rm b}}_{{\rm tr}}$. 
Similarly, suppose that we are also given two sets of unlabeled data $X^{{\rm a}}_{{\rm te}}$ and $X^{{\rm b}}_{{\rm te}}$ drawn from different marginal densities $p^{{\rm a}}_{{\rm te}} ({\bf x})$ and $p^{{\rm b}}_{{\rm te}} ({\bf x})$, respectively, in the test phase:
\begin{align}
&X^{{\rm a}}_{{\rm te}} \!=\! \{ {\bf x}^{{\rm a}} _{{\rm te}, n} \}_{n=1}^{N^{{\rm a}}_{{\rm te}}} \!\sim\! p^{{\rm a}}_{{\rm te}} ({\bf x}) \!:=\! \theta^{{\rm a}}_{{\rm te}} p^{{\rm p}}_{{\rm te}} ({\bf x}) \!+\! (1 \!-\! \theta^{{\rm a}}_{{\rm te}}) p^{{\rm n}}_{{\rm te}}  ({\bf x}), \nonumber\\
&X^{{\rm b}}_{{\rm te}} \!=\! \{ {\bf x}^{{\rm b}} _{{\rm te}, n} \}_{n=1}^{N^{{\rm b}}_{{\rm te}}} \!\sim\! p^{{\rm b}}_{{\rm te}} ({\bf x}) \!:=\! \theta^{{\rm b}}_{{\rm te}} p^{{\rm p}}_{{\rm te}} ({\bf x}) \!+\! (1 \!-\! \theta^{{\rm b}}_{{\rm te}}) p^{{\rm n}}_{{\rm te}}  ({\bf x}), \label{mar_test2}
\end{align}
where $\theta^{{\rm a}}_{{\rm te}}$ and $\theta^{{\rm b}}_{{\rm te}}$ are two class-priors such that $\theta^{{\rm a}}_{{\rm te}} \neq \theta^{{\rm b}}_{{\rm te}}$.
We assume that UU data in the test distribution are much smaller than those in the training distribution, i.e., $N _{{\rm te}}^{{\rm a}} + N _{{\rm te}}^{{\rm b}} \ll N _{{\rm tr}}^{{\rm a}} + N _{{\rm tr}}^{{\rm b}}$. 
We assume that class-priors $\pi_{{\rm te}}$, $\theta^{{\rm a}}_{{\rm tr}}$, $\theta^{{\rm b}}_{{\rm tr}}$, $\theta^{{\rm a}}_{{\rm te}}$, and $\theta^{{\rm b}}_{{\rm te}}$ are known as in previous UU learning studies
\citep{lu2018minimal,lu2020mitigating,lu2021binary}.
They can be estimated in some cases
\citep{menon2015learning,liu2015classification,jain2016estimating}.
Our goal is to learn a classifier $f$ that accurately classifies test instance ${\bf x}$ drawn from $p_{{\rm te}} ({\bf x}) = \pi_{{\rm te}} p^{{\rm p}}_{{\rm te}} ({\bf x}) + (1-\pi_{{\rm te}} ) p^{{\rm n}}_{{\rm te}} ({\bf x})$ by using unlabeled datasets $X^{{\rm a}}_{{\rm tr}} \cup X^{{\rm b}}_{{\rm tr}} \cup X^{{\rm a}}_{{\rm te}} \cup X^{{\rm b}}_{{\rm te}}$.
The proposed method can handle various learning problems, such as PN, PU, and noisy labeled learning, 
by specifying values of the class-priors as described in Section \ref{prelim}.
It can be naturally applied even when types of supervision differ between the training and test phases, e.g., noisy labeled data in the training distribution and a few PN data in the test distribution.

\subsection{Importance-weighted UU Risk}
\label{sec:uurisk}

In this subsection, we derive the importance-weighted empirical risk using UU training data by rewriting the test risk.
First, we consider the risk on the test distribution:
\begin{align}
\label{test_risk}
R_{{\rm te}} (f) := \mathbb{E} _{ p_{{\rm te}} ({\bf x}, y)} \left[ \ell (f({\bf x}), y) \right],
\end{align}
which is the objective to be minimized for learning $f$ that fits on the test distribution.
By introducing importance weight $w({\bf x},y):= p_{{\rm te}} ({\bf x}, y)/p_{{\rm tr}} ({\bf x}, y)$\footnote{
By this definition, the rewritten risk $R _{{\rm tr}}^{{\rm w}} (f)$ depends on true class-prior $\pi_{{\rm tr}}$ as shown in Eq. \eqref{iw_uu_risk1}. 
However, since the data generation process in subsection \ref{probset} does not depend on $\pi_{{\rm tr}}$, 
the value of $\pi_{{\rm tr}}$ can be freely determined in our setting (we set $\pi_{{\rm tr}}=0.5$ in our experiments). 
We use this definition to ensure consistency with existing importance weighting methods~\cite{fang2020rethinking,fang2024generalizing,kumagaiimportance}.},
as shown in previous studies
\citep{fang2020rethinking,fang2024generalizing},
this risk can be rewritten as follows,
\begin{align}
\label{iw_uu_risk1}
R_{{\rm te}} (f) &\!=\! \mathbb{E} _{ p_{{\rm tr}} ({\bf x}, y)} \left[ w({\bf x},y) \ell (f({\bf x}), y) \right] \nonumber\\ 
&\!=\! \pi _{{\rm tr}} {\mathbb E}_{p_{{\rm tr}}^{{\rm p}} ({\bf x})} \left[ \ell_w (f({\bf x}), +1) \right] \!+\! (1 \!-\! \pi_{{\rm tr}}) {\mathbb E}_{p_{{\rm tr}}^{{\rm n}} ({\bf x})} \left[ \ell_w (f({\bf x}), -1) \right] \!=:\! R _{{\rm tr}}^{{\rm w}} (f),
\end{align}
where we set $\ell_w ({\bf x}, y) := w({\bf x},y) \ell (f({\bf x}), y)$.
This reformulation shows that the test risk can be expressed as the importance-weighted risk computed on the training distribution.
While this risk depends on class-conditional densities $p ^{{\rm p}}_{{\rm tr}} ({\bf x})$ and $p_{{\rm tr}}^{{\rm n}}({\bf x} )$, by using the same procedure described in Section~\ref{prelim}, it can be rewritten with UU training densities $p^{{\rm a}}_{{\rm tr}} ({\bf x})$ and $p^{{\rm b}}_{{\rm tr}} ({\bf x})$:
\begin{align}
\label{uu_iw_risk}
R_{{\rm tr}}^{{\rm w}} (f) & = a_{{\rm tr}}  \mathbb{E}_{p_{{\rm tr}} ^{{\rm a}} ({\bf x}) } [ \ell_w(f({\bf x}), +1)] \!-\! c_{{\rm tr}}  \mathbb{E}_{p_{{\rm tr}} ^{{\rm b}} ({\bf x}) } [ \ell_w(f({\bf x}), +1)] \nonumber\\
&\!-\! b_{{\rm tr}}  \mathbb{E}_{p_{{\rm tr}} ^{{\rm a}} ({\bf x}) } [ \ell_w(f({\bf x}), -1)] \!+\! d_{{\rm tr}}  \mathbb{E}_{p_{{\rm tr}} ^{{\rm b}} ({\bf x}) } [ \ell_w(f({\bf x}), -1)],
\end{align}
where $a_{{\rm tr}} := \frac{(1-\theta^{{\rm b}}_{{\rm tr}} )\pi_{{\rm tr}} }{\theta^{{\rm a}}_{{\rm tr}} -\theta^{{\rm b}}_{{\rm tr}} }$, $b_{{\rm tr}} := \frac{\theta^{{\rm b}}_{{\rm tr} }(1-\pi_{{\rm tr}} )}{\theta^{{\rm a}}_{{\rm tr}} -\theta^{{\rm b}}_{{\rm tr}} }$, $c_{{\rm tr}}  := \frac{(1-\theta^{{\rm a}}_{{\rm tr}} )\pi_{{\rm tr}} }{\theta^{{\rm a}}_{{\rm tr}} -\theta^{{\rm b}}_{{\rm tr}} }$, and $d_{{\rm tr}}  := \frac{\theta_{{\rm tr}} ^{{\rm a}}(1-\pi_{{\rm tr}} )}{\theta_{{\rm tr}} ^{{\rm a}}-\theta_{{\rm tr}} ^{{\rm b}}}$.
Thus, the empirical estimate of Eq. \eqref{uu_iw_risk} with UU training data $X^{{\rm a}}_{{\rm tr}} \cup X^{{\rm b}}_{{\rm tr}}$ is represented as
\begin{align}
\label{uu_iw_risk_cor}
{\hat R}_{{\rm tr}}^{{\rm w}} (f) &= \left| \frac{a_{{\rm tr}}}{N_{{\rm tr}}^{{\rm a}}} \sum_{n=1}^{N_{{\rm tr}}^{{\rm a}}} \ell_w (f({\bf x} _{{\rm tr}, n}^{{\rm a}}), +1) \!-\!  \frac{c_{{\rm tr}}}{N_{{\rm tr}}^{{\rm b}}} \sum_{n=1}^{N_{{\rm tr}}^{{\rm b}}} \ell_w (f({\bf x} _{{\rm tr}, n}^{{\rm b}}), +1) \right| \nonumber\\
& \!+\! \left| \frac{d_{{\rm tr}}}{N_{{\rm tr}}^{{\rm b}}} \sum_{n=1}^{N_{{\rm tr}}^{{\rm b}}} \ell_w (f({\bf x} _{{\rm tr}, n}^{{\rm b}}), -1) \!-\! \frac{b_{{\rm tr}}}{N_{{\rm tr}}^{{\rm a}}} \sum_{n=1}^{N_{{\rm tr}}^{{\rm a}}} \ell_w (f({\bf x} _{{\rm tr}, n}^{{\rm a}}), -1)  \right|,
\end{align}
where we also used the absolute value correction to prevent overfitting.

\subsection{Importance Weight Estimation from UU Data}
\label{sec:iwe} 
Because importance weight $w({\bf x},y) = p_{{\rm te}} ({\bf x}, y)/p_{{\rm tr}} ({\bf x}, y)$ in Eq~\eqref{uu_iw_risk_cor} is not observable,
we need to estimate it with UU data.
A standard approach for this task is to use density-ratio estimation methods that directly estimate the ratio between training and test densities from data without density estimation~\citep{sugiyama2012density,kanamori2009least,bickel2009discriminative}.
However, they are often numerically unstable because
$w({\bf x},y)$ can diverge in regions where 
$p_{{\rm tr}} ({\bf x}, y)$ is small
\citep{yamada2013relative,kumagai2021meta}.
To mitigate this issue, we instead use the relative density-ratio as the importance weight:
\begin{align}
\label{relative_dre}
w_{\alpha} ({\bf x}, y) := \frac{p_{{\rm te}} ({\bf x}, y)}{\alpha p_{{\rm te}} ({\bf x}, y) + (1 - \alpha) p_{{\rm tr}} ({\bf x}, y)},
\end{align}
where $\alpha \in [0,1]$ is a hyperparameter
\citep{yamada2013relative}.
$w_{\alpha} ({\bf x}, y)$ is always bounded above by $1/\alpha$, and recovers $w({\bf x},y)$ when $\alpha=0$.
Thus, $w_{\alpha} ({\bf x}, y)$ is a bounded generalization of $w({\bf x},y)$.
The importance weighting with the relative density-ratio has shown to improve stability and performance
\citep{yamada2013relative,sakai2019covariate,kumagaiimportance}.

To estimate $w_{\alpha} ({\bf x}, y)$, we introduce model $m({\bf x}, y) \in [0,1/\alpha] $, such as a neural network.
Following previous studies
\cite{yamada2013relative,kanamori2009least,kumagaiimportance},
we determine parameters of $m({\bf x}, y)$ so that the expected squared error between true importance weight $w_{\alpha} ({\bf x}, y)$ and $m({\bf x}, y)$ is minimized:
\begin{align}
J(m) &:= \mathbb{E}_{p_{\alpha} ({\bf x}, y) } \left[ \left( w_{\alpha} ({\bf x},y) - m({\bf x}, y) \right)^2 \right] \nonumber\\
&= \mathbb{E} _{p_{{\rm te}} ({\bf x}, y)} \left[ \alpha m({\bf x}, y)^2 -2m({\bf x}, y) \right] + \mathbb{E} _{p_{{\rm tr}} ({\bf x}, y)} \left[ (1-\alpha) m({\bf x}, y)^2 \right] + C,
\end{align}
where $p_{\alpha} ({\bf x}, y) := \alpha p_{{\rm te}} ({\bf x}, y) + (1 - \alpha) p_{{\rm tr}} ({\bf x}, y)$ and $C$ is a constant term that does not depend on $m$.
Letting $M_1({\bf x},y):= \alpha m({\bf x}, y)^2 -2m({\bf x}, y)$ and $M_2({\bf x},y):= (1-\alpha) m({\bf x}, y)^2$, we have
\begin{align}
\label{iwe_j}
& J(m) \!=\! \pi_{{\rm te}} \mathbb{E} _{p_{{\rm te}}^{{\rm p}} ({\bf x})} \left[ M_1({\bf x},+1) \right] \!+\! (1 \!-\! \pi_{{\rm te}}) \mathbb{E} _{p_{{\rm te}}^{{\rm n}} ({\bf x})} \left[ M_1({\bf x},-1) \right] \nonumber\\
&\!+\! \pi_{{\rm tr}} \mathbb{E} _{p_{{\rm tr}}^{{\rm p}} ({\bf x})} \left[ M_2({\bf x},+1) \right] \!+\! (1 \!-\!\pi_{{\rm tr}} ) \mathbb{E} _{p_{{\rm tr}}^{{\rm n}} ({\bf x})} \left[ M_2({\bf x},-1) \right] + C.
\end{align}
By using Eq. \eqref{posneg_pre}, each expectation term in Eq. \eqref{iwe_j} can be rewritten as
\begin{align}
\pi_{{\rm te}} \mathbb{E} _{p_{{\rm te}}^{{\rm p}} ({\bf x})} \left[ M_1({\bf x},+1) \right] &= a_{{\rm te}} \mathbb{E} _{p_{{\rm te}}^{{\rm a}} ({\bf x})} \left[ M_1({\bf x},+1) \right]
- c_{{\rm te}} \mathbb{E} _{p_{{\rm te}}^{{\rm b}} ({\bf x})} \left[ M_1({\bf x},+1) \right], \label{m1_1} \\
(1-\pi_{{\rm te}}) \mathbb{E} _{p_{{\rm te}}^{{\rm n}} ({\bf x})} \left[ M_1({\bf x},-1) \right] &= - b_{{\rm te}} \mathbb{E} _{p_{{\rm te}}^{{\rm a}} ({\bf x})} \left[ M_1({\bf x},-1) \right]
\!+\! d_{{\rm te}} \mathbb{E} _{p_{{\rm te}}^{{\rm b}} ({\bf x})} \left[ M_1({\bf x},-1) \right], \label{m1_2} \\
\pi_{{\rm tr}} \mathbb{E} _{p_{{\rm tr}}^{{\rm p}} ({\bf x})} \left[ M_2({\bf x},+1) \right] &= a_{{\rm tr}} \mathbb{E} _{p_{{\rm tr}}^{{\rm a}} ({\bf x})} \left[ M_2({\bf x},+1) \right] - c_{{\rm tr}} \mathbb{E} _{p_{{\rm tr}}^{{\rm b}} ({\bf x})} \left[ M_2({\bf x},+1) \right], \label{m2_1} \\
(1-\pi_{{\rm tr}}) \mathbb{E} _{p_{{\rm tr}}^{{\rm n}} ({\bf x})} \left[ M_2({\bf x},-1) \right] &= - b_{{\rm tr}} \mathbb{E} _{p_{{\rm tr}}^{{\rm a}} ({\bf x})} \left[ M_2({\bf x},-1) \right]
\!+\! d_{{\rm tr}} \mathbb{E} _{p_{{\rm tr}}^{{\rm b}} ({\bf x})} \left[ M_2({\bf x},-1) \right], \label{m2_2}
\end{align}
where $a_{{\rm te}} := \frac{(1-\theta^{{\rm b}}_{{\rm te}} )\pi_{{\rm te}} }{\theta^{{\rm a}}_{{\rm te}} -\theta^{{\rm b}}_{{\rm te}} }$, $b_{{\rm te}} := \frac{\theta^{{\rm b}}_{{\rm te} }(1-\pi_{{\rm te}} )}{\theta^{{\rm a}}_{{\rm te}} -\theta^{{\rm b}}_{{\rm te}} }$, $c_{{\rm te}}  := \frac{(1-\theta^{{\rm a}}_{{\rm te}} )\pi_{{\rm te}} }{\theta^{{\rm a}}_{{\rm te}} -\theta^{{\rm b}}_{{\rm te}} }$, and $d_{{\rm te}}  := \frac{\theta_{{\rm te}} ^{{\rm a}}(1-\pi_{{\rm te}} )}{\theta_{{\rm te}} ^{{\rm a}}-\theta_{{\rm te}} ^{{\rm b}}}$.

Since $M_1({\bf x}, y) = \alpha m({\bf x}, y)^2 -2m({\bf x}, y) \geq -1/{\alpha}$ for all ${\bf x}$ and $y$, 
Eqs. \eqref{m1_1} and \eqref{m1_2} are bounded below by $-\pi_{{\rm te}}/\alpha$ and $-(1-\pi_{{\rm te}})/\alpha$, respectively. 
However, as in the risk, their empirical estimates can take smaller values and it may cause overfitting. 
To prevent this, we apply correction function $F(z) := |z-k| + k \ (k, z \in \mathbb{R} )$ where $k$ is the lower bound of the loss
\citep{ishida2020we}.
When $z \geq k$, $F(z)$ returns $z$. When $z < k$, $F(z)= 2k-z > k$. Thus, this function can penalize the loss for being smaller than $k$.
Additionally, while Eqs. \eqref{m2_1} and \eqref{m2_2} are non-negative by definition, their empirical estimates can be negative. 
We correct this by applying the absolute value function $F(z)=|z|$ as in the risk.
Consequently, the empirical estimate of Eq. \eqref{iwe_j} with the loss corrections becomes
\begin{align}
\label{obj_iw}
{\hat J}(m) &= \left| \frac{a_{{\rm te}}}{N^{{\rm a}}_{{\rm te}}} \sum_{n=1}^{N^{{\rm a}}_{{\rm te}}} M_1({\bf x}^{{\rm a}}_{{\rm te}, n}, +1) - \frac{c_{{\rm te}}}{N^{{\rm b}}_{{\rm te}}} \sum_{n=1}^{N^{{\rm b}}_{{\rm te}}} M_1({\bf x}^{{\rm b}}_{{\rm te}, n}, +1) - k_1 \right| \nonumber\\
&+ \left| \frac{d_{{\rm te}}}{N^{{\rm b}}_{{\rm te}}} \sum_{n=1}^{N^{{\rm b}}_{{\rm te}}} M_1({\bf x}^{{\rm b}}_{{\rm te}, n}, -1) - \frac{b_{{\rm te}}}{N^{{\rm a}}_{{\rm te}}} \sum_{n=1}^{N^{{\rm a}}_{{\rm te}}} M_1({\bf x}^{{\rm a}}_{{\rm te}, n}, -1) - k_2 \right| \nonumber\\
&+ \left| \frac{a_{{\rm tr}}}{N^{{\rm a}}_{{\rm tr}}} \sum_{n=1}^{N^{{\rm a}}_{{\rm tr}}} M_2({\bf x}^{{\rm a}}_{{\rm tr}, n}, +1) - \frac{c_{{\rm tr}}}{N^{{\rm b}}_{{\rm tr}}} \sum_{n=1}^{N^{{\rm b}}_{{\rm tr}}} M_2({\bf x}^{{\rm b}}_{{\rm tr}, n}, +1) \right| \nonumber\\
&+ \left| \frac{d_{{\rm tr}}}{N^{{\rm b}}_{{\rm tr}}} \sum_{n=1}^{N^{{\rm b}}_{{\rm tr}}} M_2({\bf x}^{{\rm b}}_{{\rm tr}, n}, -1) - \frac{b_{{\rm tr}}}{N^{{\rm a}}_{{\rm tr}}} \sum_{n=1}^{N^{{\rm a}}_{{\rm tr}}} M_2({\bf x}^{{\rm a}}_{{\rm tr}, n}, -1) \right|,
\end{align}
where $k_1 := - \pi_{{\rm te}}/\alpha$, $k_2 := - (1 - \pi_{{\rm te}})/\alpha $, and we omit constant terms that do not depend on $m$.

\subsection{Classifier Loss Function}
\label{sec:cl}

Our loss function for learning classifier $f$ is a weighted sum of the empirical UU risk on the test distribution and the importance-weighted empirical UU risk on the training distribution in Eq. \eqref{uu_iw_risk_cor}:
\begin{align}
\label{our_loss}
{\hat L} (f, m) := \beta {\hat R} _{{\rm te}} (f) + (1-\beta) {\hat R} _{{\rm tr}}^{{\rm w}} (f, m),
\end{align}
where $\beta \in [0,1)$ is a weighting hyperparameter, and $ {\hat R} _{{\rm te}} (f)$ is the empirical test UU risk calculated with $X^{{\rm a}}_{{\rm te}} \cup X^{{\rm b}}_{{\rm te}}$, 
\begin{align}
\label{em_uu_te}
{\hat R} _{{\rm te}} (f) &= \left| \frac{a_{{\rm te}}}{N_{{\rm te}}^{{\rm a}}} \sum_{n=1}^{N_{{\rm te}}^{{\rm a}}} \ell (f({\bf x} _{{\rm te}, n}^{{\rm a}}), +1) - \frac{c_{{\rm te}}}{N_{{\rm te}}^{{\rm b}}} \sum_{n=1}^{N_{{\rm te}}^{{\rm b}}} \ell (f({\bf x} _{{\rm te}, n}^{{\rm b}}), +1) \right| \nonumber\\
&+ \left| \frac{d_{{\rm te}}}{N_{{\rm te}}^{{\rm b}}} \sum_{n=1}^{N_{{\rm te}}^{{\rm b}}} \ell (f({\bf x} _{{\rm te}, n}^{{\rm b}}), -1) - \frac{b_{{\rm te}}}{N_{{\rm te}}^{{\rm a}}} \sum_{n=1}^{N_{{\rm te}}^{{\rm a}}} \ell (f({\bf x} _{{\rm te}, n}^{{\rm a}}), -1)  \right|.
\end{align}
Here, we use Eq. \eqref{uu_risk_cor} to derive Eq. \eqref{em_uu_te} and explicitly describe the dependency of the importance weight model $m$ in ${\hat R} _{{\rm tr}}^{{\rm w}}$ for clarity. 
Although unlabeled data $X^{{\rm a}}_{{\rm te}} \cup X^{{\rm b}}_{{\rm te}}$ are used for estimating the importance weights, they are not directly used for learning $f$ in ${\hat R} _{{\rm tr}}^{{\rm w}}$.
By using both empirical risks, we can use all available data $X^{{\rm a}}_{{\rm tr}} \cup X^{{\rm b}}_{{\rm tr}} \cup X^{{\rm a}}_{{\rm te}} \cup X^{{\rm b}}_{{\rm te}}$ for learning $f$ directly.

\subsection{Training}
\label{train_pro}

To compute ${\hat R} _{{\rm tr}}^{{\rm w}} (f, m)$, the conventional importance weighting approach first estimates the importance weights and 
subsequently uses them to calculate the importance-weighted risk
\citep{yamada2013relative,sugiyama2012density,kanamori2009least,bickel2009discriminative}.
However,
the importance weight is difficult to estimate when
using complex models such as neural networks or complex data such as image data 
\cite{fang2020rethinking,kumagaiimportance,kato2021non,rhodes2020telescoping}.
Because the estimation error directly affects the subsequent classifier learning, 
this two-step approach often does not work well
\cite{fang2020rethinking,kumagaiimportance}.
To alleviate this, recent methods use a dynamic approach that iterates between the importance weight estimation and classifier learning while sharing a neural network for feature extraction
\cite{fang2020rethinking,fang2024generalizing,kumagaiimportance}.
By sharing the feature extractor, they can effectively perform the importance weighting.
\begin{algorithm}[t]
\caption{Training procedure of the proposed method}
\label{arg}
\begin{algorithmic}[1]
\REQUIRE UU data in the training and test distributions $X^{{\rm a}}_{{\rm tr}} \cup X^{{\rm b}}_{{\rm tr}} \cup X^{{\rm a}}_{{\rm te}} \cup X^{{\rm b}}_{{\rm te}}$, mini-batch sizes for the training and test distributions $(B_{{\rm tr}}, B_{{\rm te}})$, 
positive class-priors $(\pi_{{\rm tr}}, \pi_{{\rm te}}, \theta^a_{{\rm tr}}, \theta^b_{{\rm tr}}, \theta^a_{{\rm te}}, \theta^b_{{\rm te}})$,
relative parameter $\alpha$, and weighting parameter $\beta$. 
\ENSURE Parameters of neural networks $h$, $u$, and $v$.
\REPEAT
\STATE{Sample UU data with size $B_{{\rm tr}} $ from $X^{{\rm a}}_{{\rm tr}} \cup X^{{\rm b}}_{{\rm tr}}$}
\STATE{Sample UU data with size $B_{{\rm te}} $ from $X^{{\rm a}}_{{\rm te}} \cup X^{{\rm b}}_{{\rm te}}$}

\STATE \textbf{\# Importance weight estimation}
\STATE{Calculate loss in Eq. \eqref{obj_iw} on the sampled data} 

\STATE{Update parameters of $v$ with the gradient of the loss fixing feature extractor $h$}

\STATE \textbf{\# Classifier learning}
\STATE{Calculate the loss in Eq. \eqref{our_loss} on the sampled data with current importance weights}
\STATE{Update classifier parameters of $u$ and $h$ with the gradient of the loss fixing the importance weights}
\UNTIL{End condition is satisfied;}
\end{algorithmic}
\end{algorithm}

The proposed method also follows this dynamic approach.
Specifically, we use the following neural networks for modeling classifiers and importance weights,
\begin{align}
f({\bf x}) := u(h({\bf x})), \ m({\bf x}, y) := v([h({\bf x}), y]),
\end{align}
where $h: {\cal X} \to \mathbb{R}^K$, $u: \mathbb{R}^K \to \mathbb{R}$, and $v: \mathbb{R}^{K+2} \to \mathbb{R}$ are neural networks for feature extraction, classifiers, and importance weights, respectively. Here, we assume that label $y \in \{-1,+1\}$ is represented as one-hot encoding and $[\cdot, \cdot]$ is a concatenation.
Algorithm \ref{arg} shows the training procedure of our method with stochastic gradient descent methods.
We first randomly sample the mini-batch data (UU data) from the training and test distributions (Lines 2--3).
Then, we calculate the loss in Eq. \eqref{obj_iw} for the importance weight estimation (Line 5) and update parameters of $v$ with the gradient of the loss fixing feature extractor $h$ (Line 6).
We fixed $h$ to avoid overfitting as in 
\citep{fang2020rethinking,kumagaiimportance}.
Then, by using the estimated importance weights, we calculate the loss in Eq. \eqref{our_loss} for classifier learning (Line 8).
We update parameters of classifier $u$ and $h$ by using the gradient of the loss (Line 9).
In this step, we fixed the importance weights to avoid learning a meaningless model, $m({\bf x}, y)=0$ for all ${\bf x}$ and $y$.

\section{Experiments}

\subsection{Data}
\label{exp_data}
We used four real-world datasets in the main paper: 
MNIST
\citep{lecun1998gradient}, 
FashionMNIST (FMNIST)
\citep{xiao2017fashion}, 
CIFAR10~\citep{krizhevsky2009learning},
and DIABETES
\citep{gardner2024benchmarking}.
They have been commonly used in recent distribution shift adaptation studies~\cite{kumagaiimportance,kumagai2024auc,fang2020rethinking,qian2024efficient}.
MNIST and FMNIST are $28 \times 28$ grayscale image datasets, while CIFAR10 consists of $32 \times 32$ RGB images. DIABETES is a tabular dataset with distribution shift \citep{gardner2024benchmarking}, where the task is to predict whether a respondent has diabetes from $142$ dimensional features. 
Data from the training and test distributions are ``white non-hispanic'' and other racial groups, respectively.

For MNIST, FMNIST, and CIFAR10, following the previous studies~\cite{kumagaiimportance,fang2024generalizing}, we constructed binary classification tasks by partitioning the original classes of each dataset into positive and negative classes, and created two types of distribution shifts:
the first one is the support shift,
where the support of input ${\bf x}$ changes between the training and testing phases.
This shift is a particularly challenging type of covariate shift~\cite{fang2024generalizing,kumagaiimportance}.
The second one is the input-output relation shift (concept shift),
where the inputs' support does not change but $p(y| {\bf x})$ varies between the training and testing phases.
Note that no methods use the knowledge that such shifts have occurred in our experiments.
The detailed construction procedure is provided in Section~\ref{dc_cs}.

For UU data in the training distribution $X^{{\rm a}}_{{\rm tr}}$ and $X^{{\rm b}}_{{\rm tr}}$,
we set the numbers of the instances $(N_{{\rm tr}}^a, N_{{\rm tr}}^b)$ to $(2500, 2500)$.
For UU data in the test distribution $X^{{\rm a}}_{{\rm te}}$ and $X^{{\rm b}}_{{\rm te}}$,
we changed the numbers of the instances $(N_{{\rm te}}^a, N_{{\rm te}}^b)$ within $\{(50,50), (100, 100), (150,150)\}$.
For these UU datasets,
we changed the class-priors $(\theta^{{\rm a}},\theta^{{\rm b}}):=(\theta^{{\rm a}}_{{\rm tr}}, \theta^{{\rm b}}_{{\rm tr}}) = (\theta^{{\rm a}}_{{\rm te}}, \theta^{{\rm b}}_{{\rm te}})$ within $\{(0.8, 0.2), (0.7, 0.3), (0.6, 0.4)\}$ as in the previous studies
\cite{lu2020mitigating}.
We also used different UU data in the test distribution where each size is $100$ as validation data.
We used $2000$ positive and $2000$ negative data in the test distribution as test data for evaluation (i.e., $\pi_{{\rm te}} = 0.5$). 
Training, validation, and test datasets did not overlap.
We conducted 10 experiments for each pair of the number of UU data in the test distribution and the class-priors while changing the random seeds and evaluated the mean test accuracy. 

\subsection{Comparison Methods}

We compared our method with five UU learning-based methods:
UU learning method on the test distribution (teUU)
\cite{lu2020mitigating},
UU learning method on the training distribution (trUU),
multi-task learning method of UU learning (mtUU),
multi-task learning method of UU learning with a single neural network (mtsUU),
and
domain adaptation method for UU learning (daUU).
All methods including our method used neural networks for classifiers and the absolute value correction to prevent overfitting.

teUU and trUU learn the classifier by minimizing the non-negative empirical risk (in Eq. \eqref{uu_risk_cor}) with UU data in the test and training distributions, respectively.
Note that teUU and trUU are state-of-the-art UU learning methods.
Since no distribution adaptation methods for UU learning exist, we created several adaptation methods (mtUU, mtsUU, and daUU).
mtUU and mtsUU learn the classifier by minimizing the weighted sum of the non-negative empirical UU risks on the training and test distributions.
mtUU used a two-head neural network for handling the difference between the training and test distributions.
mtsUU used the single neural network as in the proposed method.
The difference between the proposed method and mtsUU is whether or not the importance weights are used in Eq. \eqref{our_loss} (i.e., mtsUU used $m({\bf x},y)=1$ for all ${\bf x}, y$).
daUU minimizes the weighted sum of the non-negative empirical UU risks on both distributions (i.e., the loss of mtsUU) 
while minimizing the feature discrepancy 
to obtain invariant features as in
\citep{sonntag2022positive}.
We used the maximum mean discrepancy (MMD) loss
\citep{li2015generative} 
to minimize the feature discrepancy, which is commonly used for distribution shift adaptation
\citep{farahani2021brief,kumagaiimportance}.
The details of network architectures and hyperparameters are described in Sections \ref{arc} and \ref{hyp}. 

\subsection{Results}

\begin{table*}[t]
\caption{Average test accuracies over different UU test data sizes $(N_{{\rm te}}^{{\rm a}}, N_{{\rm te}}^{{\rm b}})$ within $ \{ (50, 50), (100, 100), (150, 150) \} $ and class-priors $(\theta^{{\rm a}},\theta^{{\rm b}})$ within $\{(0.8, 0.2), (0.7, 0.3), (0.6, 0.4)\}$.
In the Shift column, `S' and `IO' represent the support shift and input-output relation shift, respectively.
Values in bold are not statistically different at the $5\%$ level from the best performing method in each row according to a paired t-test.
}
\label{result_all}
\centering
\scalebox{0.95}{
\begin{tabular}{llrrrrrrrr}
\hline
Data & Shift & \multicolumn{1}{c}{Ours} & \multicolumn{1}{c}{teUU} & \multicolumn{1}{c}{trUU} & \multicolumn{1}{c}{mtsUU} &  \multicolumn{1}{c}{mtUU} &
\multicolumn{1}{c}{daUU} \\
\hline
MNIST & S & \bf{0.7482} & \bf{0.7419} & 0.5879 & 0.7388 & \bf{0.7355} & 0.7423 \\
 & IO & 0.7211 & \bf{0.7334} & 0.5412 & 0.6922 & 0.6996 & 0.6929  \\
\hline
FMNIST & S & \bf{0.9620} & 0.9531 & 0.9161 & 0.9551 & 0.9509 & 0.9550 \\
 & IO & \bf{0.8841} & \bf{0.8929} & 0.7278 & 0.8796 & 0.8823 & 0.8769 \\
\hline
CIFAR10 & S & \bf{0.8294} & 0.6929 & 0.8258 & 0.8281 & 0.8166 & \bf{0.8308} \\
 & IO & \bf{0.7125} & 0.6864 & 0.6665 & \bf{0.7084} & 0.6947 & \bf{0.7099} \\
\hline
DIABETES & & \bf{0.7090} & 0.6300 & \bf{0.7115} & \bf{0.7117} & 0.7011 & \bf{0.7118} \\
\hline
\end{tabular}
}
\end{table*}

\begin{table*}[t!]
\caption{Results with a few PN data in the test distribution: Average test accuracies over different UU test data sizes $(N_{{\rm te}}^{{\rm a}}, N_{{\rm te}}^{{\rm b}})$ within $ \{ (10,10), (50, 50), (100, 100) \} $ and training class-priors $(\theta_{{\rm tr}}^{{\rm a}},\theta_{{\rm tr}}^{{\rm b}})$ within $\{(0.8, 0.2), (0.7, 0.3), (0.6, 0.4)\}$. Class-priors of UU test data $(\theta_{{\rm te}}^{{\rm a}},\theta_{{\rm te}}^{{\rm b}})$ were set to $(1,0)$.
}
\label{result_all_diftheta2}
\centering
\scalebox{0.95}{
\begin{tabular}{llrrrrrrrr}
\hline
Data & Shift & \multicolumn{1}{c}{Ours} & \multicolumn{1}{c}{teUU} & \multicolumn{1}{c}{trUU} & \multicolumn{1}{c}{mtsUU} &  \multicolumn{1}{c}{mtUU} &
\multicolumn{1}{c}{daUU} \\
\hline
MNIST & S & 0.8237 & 0.8184 & 0.6012 & 0.8194 & \bf{0.8309} & 0.8177  \\
 & IO & \bf{0.8075} & \bf{0.8102} & 0.5543 & 0.8012 & \bf{0.8127} & 0.7998  \\
\hline
FMNIST & S & \bf{0.9726} & 0.9655 & 0.9181 & \bf{0.9719} & 0.9705 & \bf{0.9717}  \\
 & IO & \bf{0.9367} & 0.9285 & 0.7505 & 0.9364 & 0.9299 & \bf{0.9393}  \\
\hline
CIFAR10 & S & \bf{0.8492} & 0.7555 & 0.8286 & 0.8422 & 0.8376 & \bf{0.8466} \\
 & IO & \bf{0.7605} & 0.7455 & 0.6732 & 0.7563 & 0.7492 & \bf{0.7611}  \\
\hline
DIABETES & & \bf{0.7251} & 0.6757 & \bf{0.7233} & \bf{0.7252} & 0.7209 & \bf{0.7238}  \\
\hline
\end{tabular}
}
\end{table*}

Table \ref{result_all} shows the average test accuracies for each dataset. 
The full results with the standard deviations are described in subsection \ref{full_dev}.
The proposed method performed the best or comparably to it in almost all cases (6 out of 7 cases).
teUU and trUU tended to perform worse than the proposed method because teUU used only a small amount of UU data in the test distribution and trUU used only UU data in the training distribution.
Although mtsUU and mtUU used UU data in both distributions as in the proposed method,
their performance was limited since they do not have an explicit mechanism for transfer.
As mtsUU is the proposed method without importance weights, the result highlights the effectiveness of the importance weights.
Although daUU aims to transfer by learning invariant features, it does not explicitly target the test risk. In contrast, the proposed method achieved better performance in more cases by directly minimizing the test risk within the importance-weighting framework.
The results with each number of UU data in the test distribution and with each class-priors were described in subsections \ref{dif_ndata} and \ref{dif_cps}, respectively. 
The proposed method worked well on each case.
\begin{wrapfigure}[11]{r}{47mm}
  \centering
  \includegraphics[width=5.2cm]{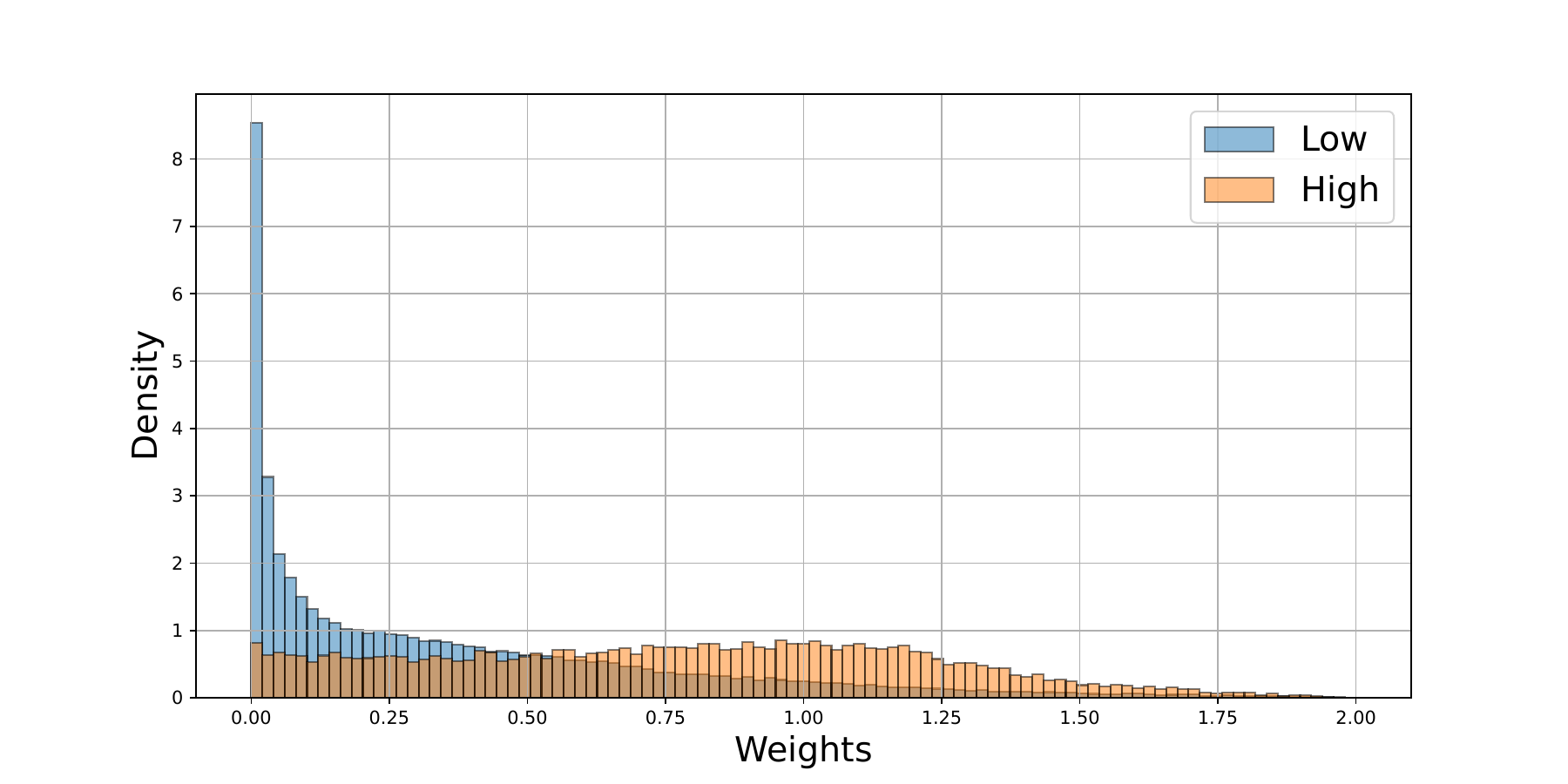}
  \caption{Importance weight distribution of the proposed method on MNIST with the support shift.}
  \label{fig:iwv}
\end{wrapfigure}

Figure \ref{fig:iwv} shows the distribution of estimated importance weights of the proposed method on MNIST dataset with the support shift when 
$(N_{{\rm te}}^{{\rm a}}, N_{{\rm te}}^{{\rm b}})=(50, 50)$ and $(\theta^{{\rm a}},\theta^{{\rm b}})=(0.7, 0.3)$.
‘High' and ‘Low' represent data in the test distribution and data that are not in the test distribution, respectively.
The proposed method was able to correctly estimate importance weights, i.e., the weights of ‘High' and ‘Low' are large and small, respectively.

Table \ref{result_all_diftheta2} shows the results when types of supervision were different between the training and testing phases:
a few PN data in the test distribution and noisy labeled data in the training distribution.
We evaluated this setting because a few data might be accurately labeled by domain experts.
Considering the cost of precise labeling, we used fewer instances than in previous setting.
The proposed method also performed the best or comparably to it in 6 out of 7 cases.
This result shows that the proposed method works well even with different types of supervision.

\section{Conclusion}

In this paper, we proposed a distribution shift adaptation method for UU learning with the importance weighting.
Our method can handle various types of learning problems under distribution shift such as PN, PU, noisy label, and similarity-based learning. 
Moreover, it can perform adaptation without assumptions of the shift types.
The experiments showed the effectiveness of our method.

\bibliographystyle{abbrv}
\bibliography{ref_uu}


\appendix

\section{Construction Procedure of Support and Input-output Relation Shifts}
\label{dc_cs}

In this section, we describe the construction procedure for the distribution shifts, which is the same as that used in the previous study \citep{kumagaiimportance}.

We first describe the procedure to construct the support shift.
For MNIST, even and odd digits were treated as negative and positive, respectively.
We used digits 0, 2, and 4 (4, 6, and 8) for negative- and digits 1, 3, and 5 (5, 7, and 9) for positive-conditional densities of the training (test) distribution.
For FMNIST, the upper garments (0, 2, 3, 4, and 6) and the others were treated as negative and positive, respectively, where numbers in parentheses represent class labels.
We used class labels 0, 2, and 3 (3, 4, and 6) for negative- and class labels 1, 5, and 7 (7, 8, and 9) for positive-conditional densities of the training (test) distribution.
For CIFAR10, the animal categories (2, 3, 4, 5, 6, and 7) and the others were treated as negative and positive, respectively.
We used class labels 2, 3, 4, and 5 (4, 5, 6, and 7) as negative and class labels 0, 1, and 8 (1, 8, and 9) as positive in the training (test) distribution.

We next describe the procedure to construct the input-output shift.
For MNIST, we used even digits for negative- and odd digits for positive-conditional densities of the test distribution.
For the training distribution,
we swapped the digits 0 and 2 with 1 and 3 (e.g., the digits 0, 2, 5, 7, and 9 were used for the positive-conditional density of the training distribution).
For FMNIST,
we used upper garments (0, 2, 3, 4, and 6) for negative- and the others for positive-conditional densities of the test distribution.
We swapped the class labels 0 and 2 with 1 and 5 in the training distribution.
For CIFAR10, we used the animal categories (2, 3, 4, 5, 6, and 7) for negative- and the vehicles for positive-conditional densities of the test distribution.
We swapped the class labels 2 and 3 with 0 and 1 in the training distribution.

\section{Neural Network Architectures}
\label{arc}

For MNIST, FMNIST, and DIABETES, a three-layered feed-forward neural network with ReLU activation was used for feature extractor $h$.
The number of hidden and output nodes was $128$.
For CIFAR10, a convolutional neural network, which consisted of two convolutional blocks followed by a two-layered feed-forward neural network, was used for feature extractor $h$. The first (second) convolutional block comprised a 6 (16) filter $5 \times 5$ convolution, the ReLU activation, and a $2 \times 2$ max-pooling layer. The numbers of the hidden and output nodes were 120 and 84, respectively.
One-layered and two-layered feed-forward neural networks were used for $u$ and $v$, respectively.
For the output activation of $v$, we used $\frac{1}{\alpha} \sigma(\cdot)$, where $\sigma$ is a sigmoid function, to match the value range of the relative density-ratio (importance weights). For all comparison methods, the same neural network architecture (i.e., $u(h({\bf x}))$) was used for the classifier.
For mtUU, two different classifier heads (i.e., $u$) were used for the training and test distributions.

\section{Hyperparameters}
\label{hyp}

For all methods, the empirical UU risk with validation data on the test distribution was used to select hyperparameters and early-stopping to mitigate overfitting.
For all methods, the sigmoid loss was used for loss $\ell$ as in the previous study
\citep{kiryo2017positive,kumagaiimportance}.
For the proposed method, relative parameter $\alpha$ was selected from $\{0.1, 0.5, 0.9\}$ and
training class-prior $\pi_{{\rm tr}}$ was set to $0.5$.
For the proposed method, mtUU, mtsUU, and daUU, weighting parameter $\beta$ was selected from $\{0, 0.01, 0.1, 0.3, 0.5, 0.7, 0.9 \}$.
Note that the proposed method, mtUU, and mtsUU are equivalent to teUU when $\beta=1$.
For daUU, the MMD loss was applied to $h({\bf x})$. The number of RBF kernel mixtures was set to $5$ as in the previous study
\citep{li2015generative}.
The weighting parameter of the MMD loss was chosen from $\{10^{-1}, 10^{-2}, 10^{-3} \}$.
The mini-batch sizes for the training and test distributions, $B_{{\rm tr}}$ and $B_{{\rm te}}$, were set to $512$.
For all methods, we used the Adam optimizer
\citep{kingma2014adam}.
We set the learning rate to $10^{-4}$ and $10^{-3}$ for classifier learning and importance weight estimation, respectively.
The maximum number of epochs was $200$.
All methods were implemented using Pytorch
\citep{paszke2017automatic},
and all experiments were conducted on a Linux server with an Intel Xeon CPU and A100 GPU.

\begin{figure}[t]
    \centering
    \begin{minipage}{0.24\linewidth}
        \centering
        \includegraphics[width=3.4cm]{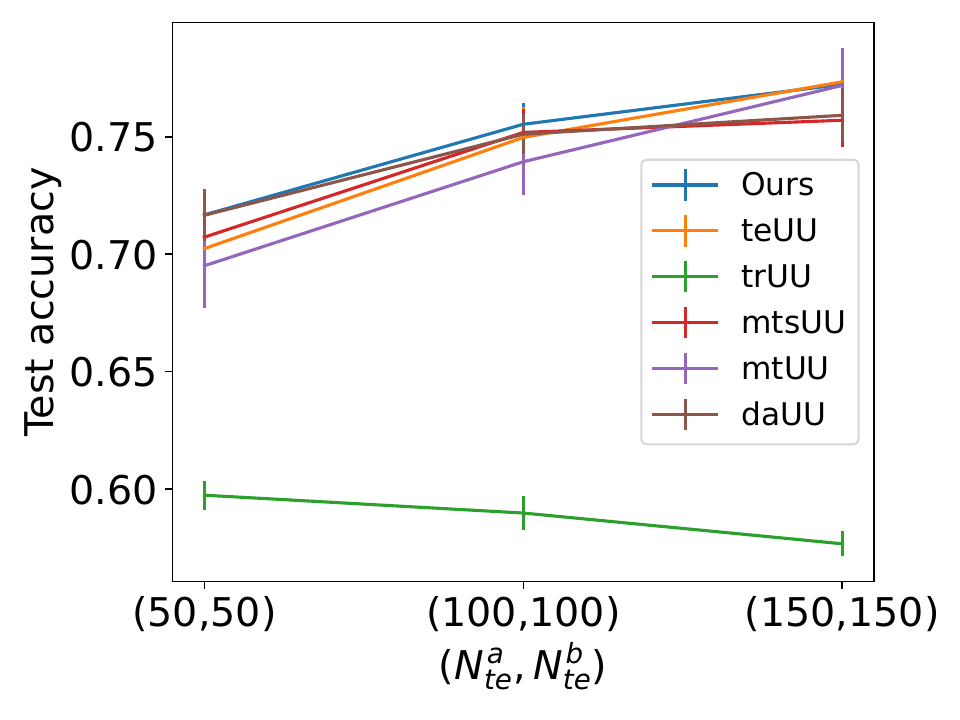}
        \subcaption{MNIST(S)}
    \end{minipage}
    \hfill
    \begin{minipage}{0.24\linewidth}
        \centering
        \includegraphics[width=3.4cm]{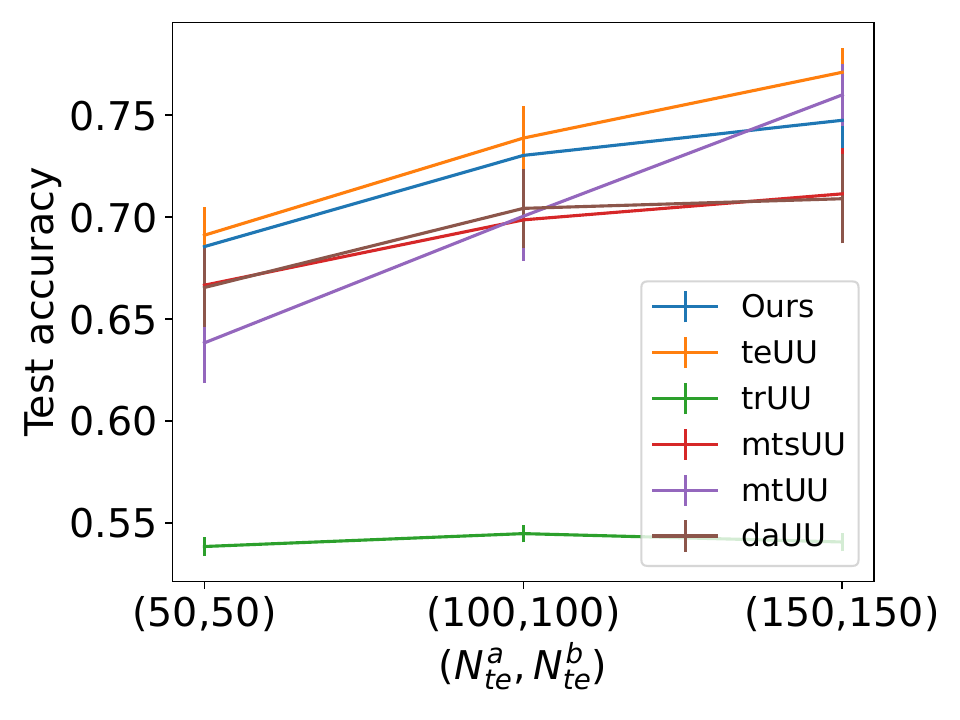}
        \subcaption{MNIST(IO)}
    \end{minipage}
    \hfill
    \begin{minipage}{0.24\linewidth}
        \centering
        \includegraphics[width=3.4cm]{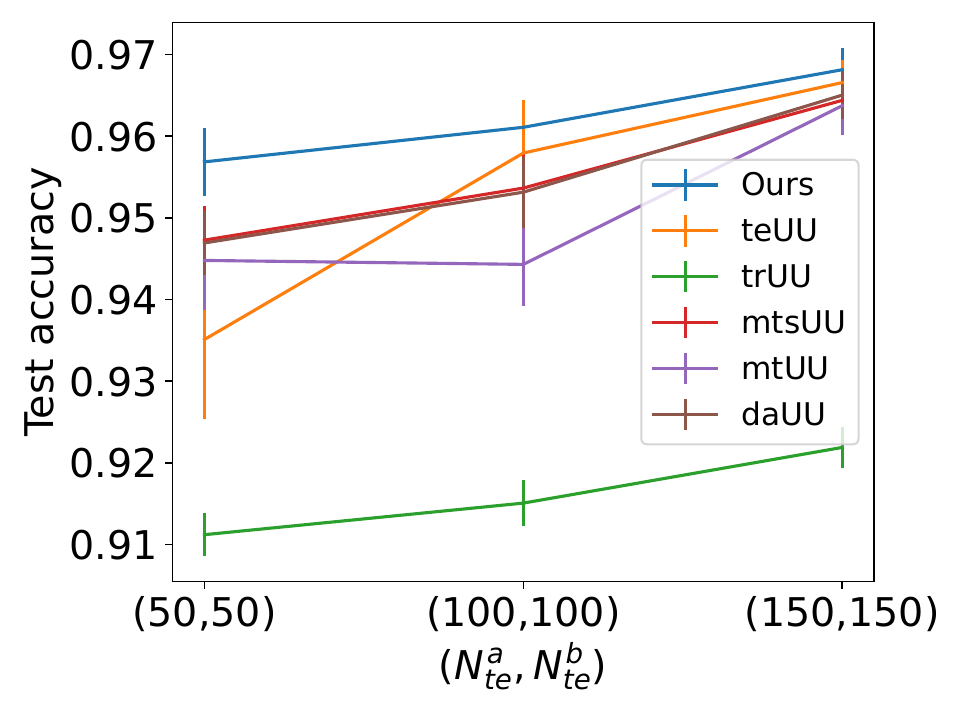}
        \subcaption{FMNIST(S)}
    \end{minipage}
    \hfill
    \begin{minipage}{0.24\linewidth}
        \centering
        \includegraphics[width=3.4cm]{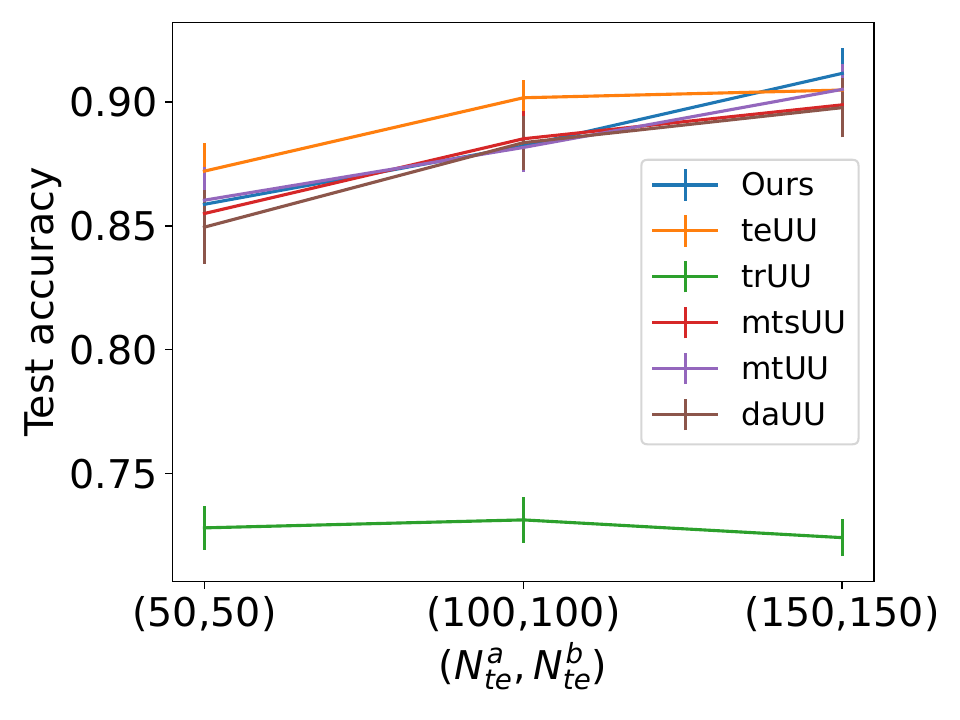}
        \subcaption{FMNIST(IO)}
    \end{minipage}
    \hfill
    \begin{minipage}{0.3\linewidth}
        \centering
        \includegraphics[width=3.4cm]{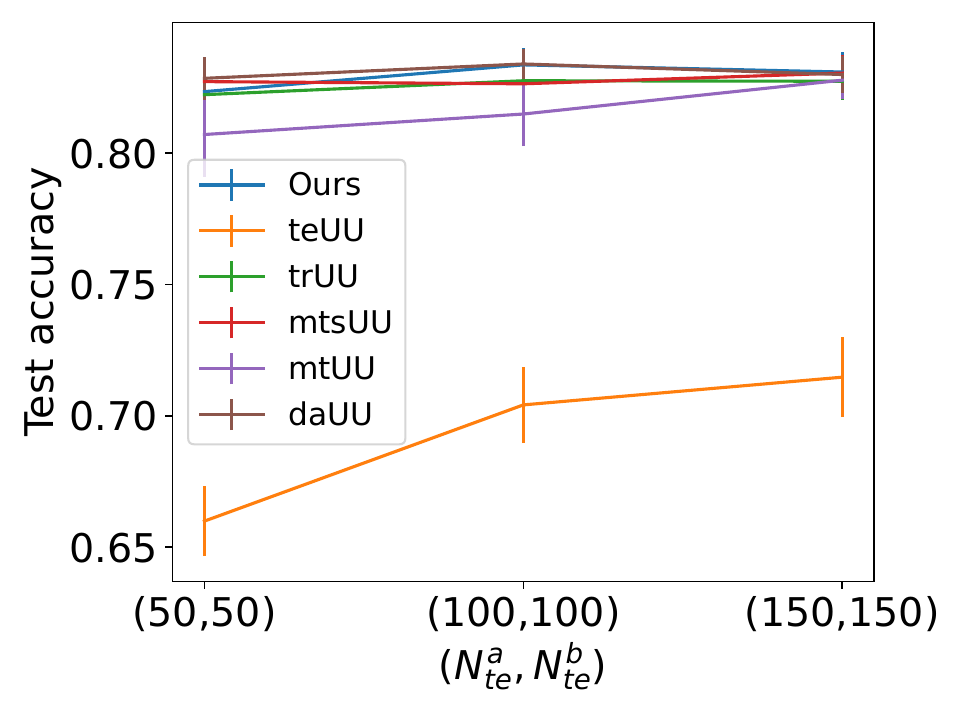}
        \subcaption{CIFAR10(S)}
    \end{minipage}
    \begin{minipage}{0.3\linewidth}
        \centering
        \includegraphics[width=3.4cm]{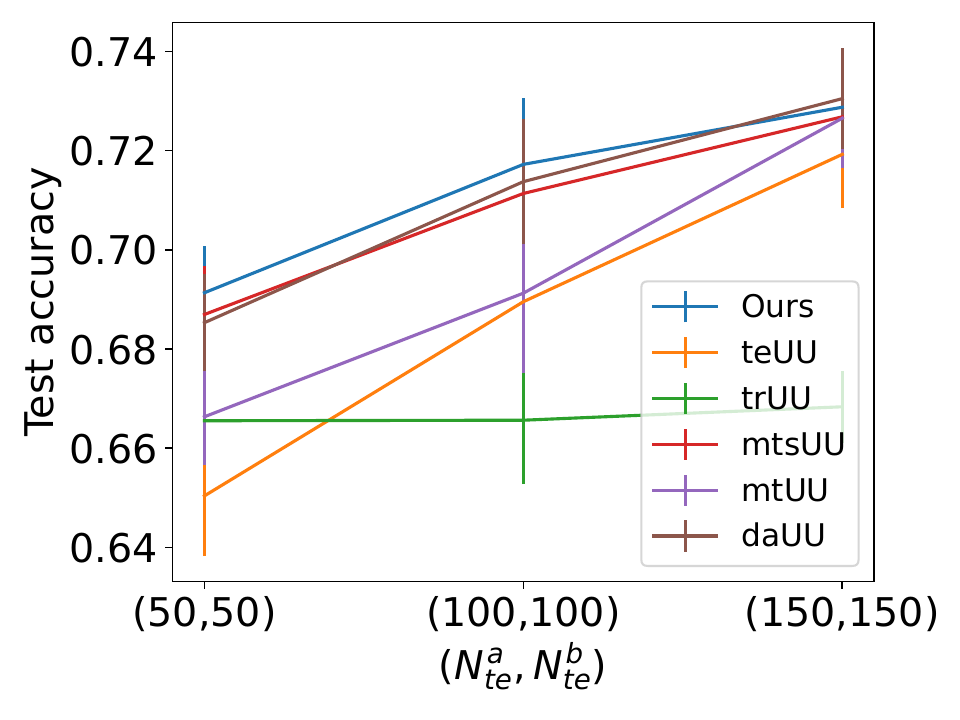}
        \subcaption{CIFAR10(IO)}
    \end{minipage}
    \begin{minipage}{0.3\linewidth}
        \centering
        \includegraphics[width=3.4cm]{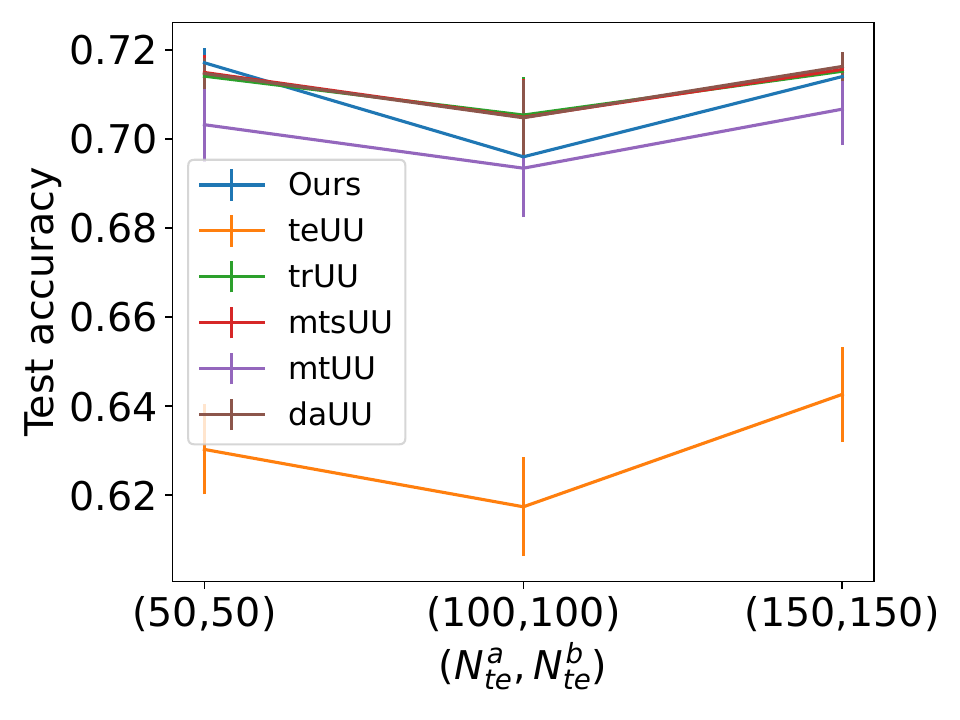}
        \subcaption{DIABETES}
    \end{minipage}
    \caption{The average test accuracies with the standard errors when changing the number of UU data in the test distribution.}
    \label{result_ndata}
\end{figure}

\section{Additional Experimental Results}
\label{appenres}

\subsection{Results with Different Numbers of UU Data in the Test Distribution}
\label{dif_ndata}

Figure \ref{result_ndata} shows the average test accuracies with the standard errors when changing the numbers of UU data
in the test distribution in the case where data of both the training and test distributions are noisy.
As the number of UU data increased, the performance of the proposed method improved as expected.
The proposed method tended to work well in each UU data size.

\begin{figure}[t]
    \centering
    \begin{minipage}{0.24\linewidth}
        \centering
        \includegraphics[width=3.4cm]{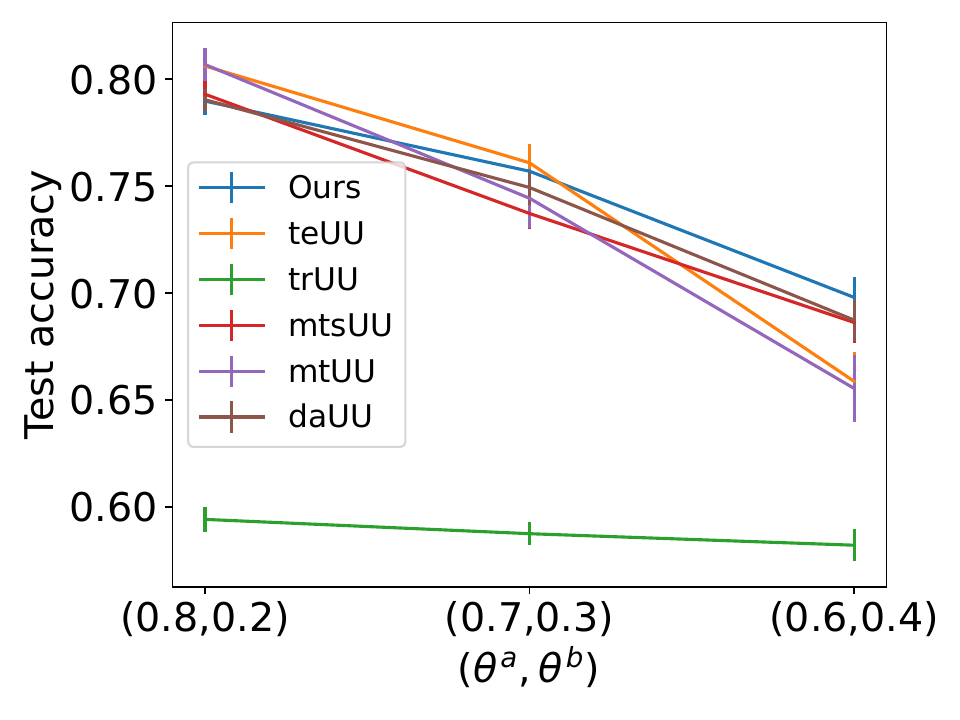}
        \subcaption{MNIST(S)}
    \end{minipage}
    \hfill
    \begin{minipage}{0.24\linewidth}
        \centering
        \includegraphics[width=3.4cm]{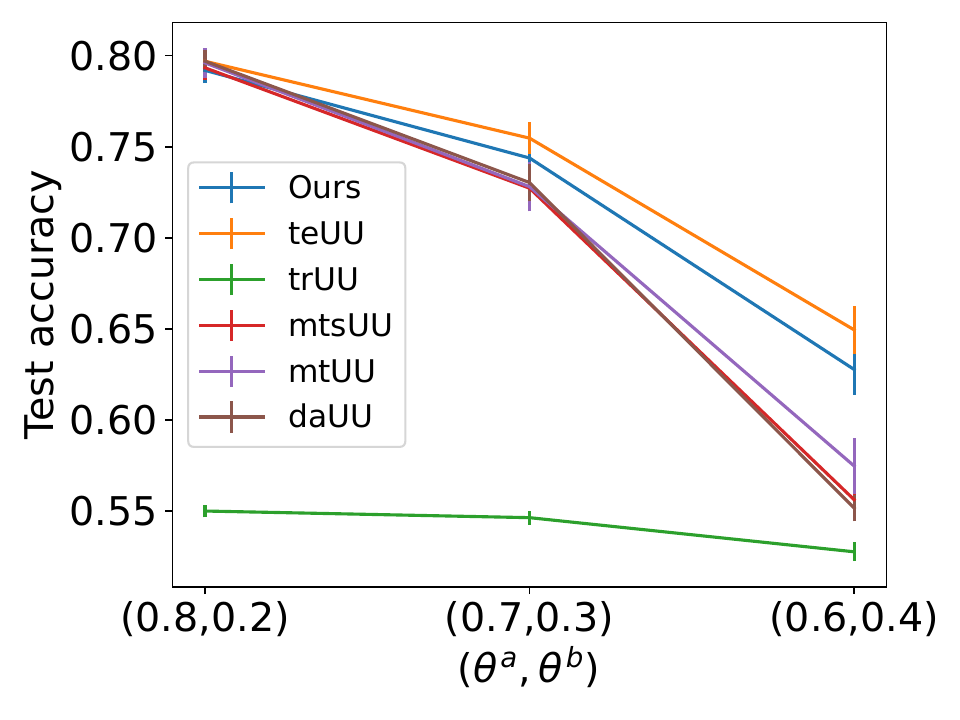}
        \subcaption{MNIST(IO)}
    \end{minipage}
    \hfill
    \begin{minipage}{0.24\linewidth}
        \centering
        \includegraphics[width=3.4cm]{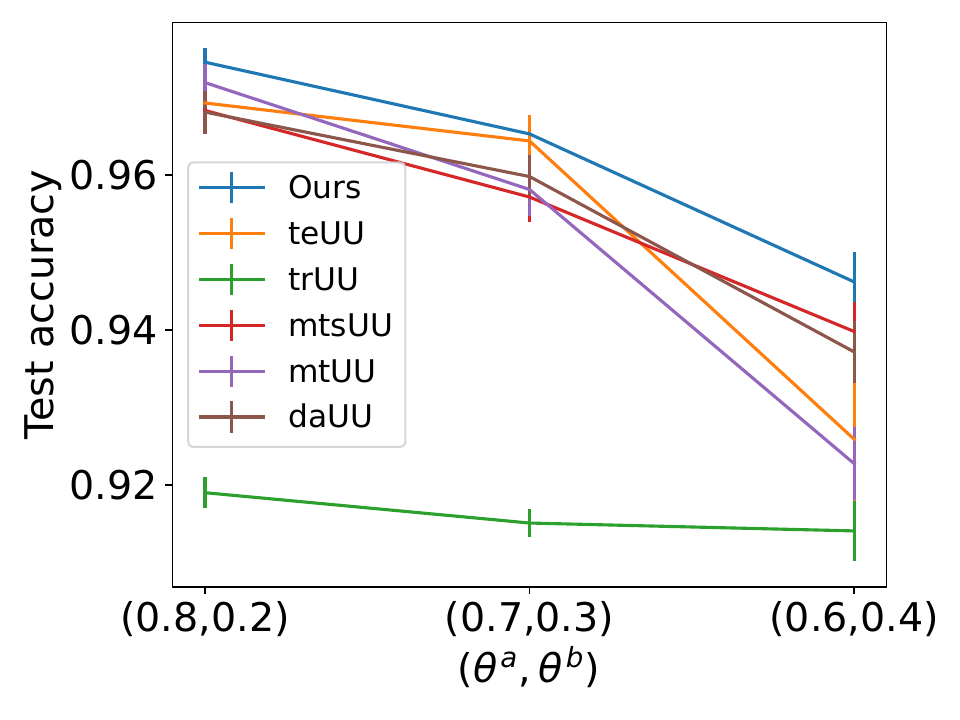}
        \subcaption{FMNIST(S)}
    \end{minipage}
    \hfill
    \begin{minipage}{0.24\linewidth}
        \centering
        \includegraphics[width=3.4cm]{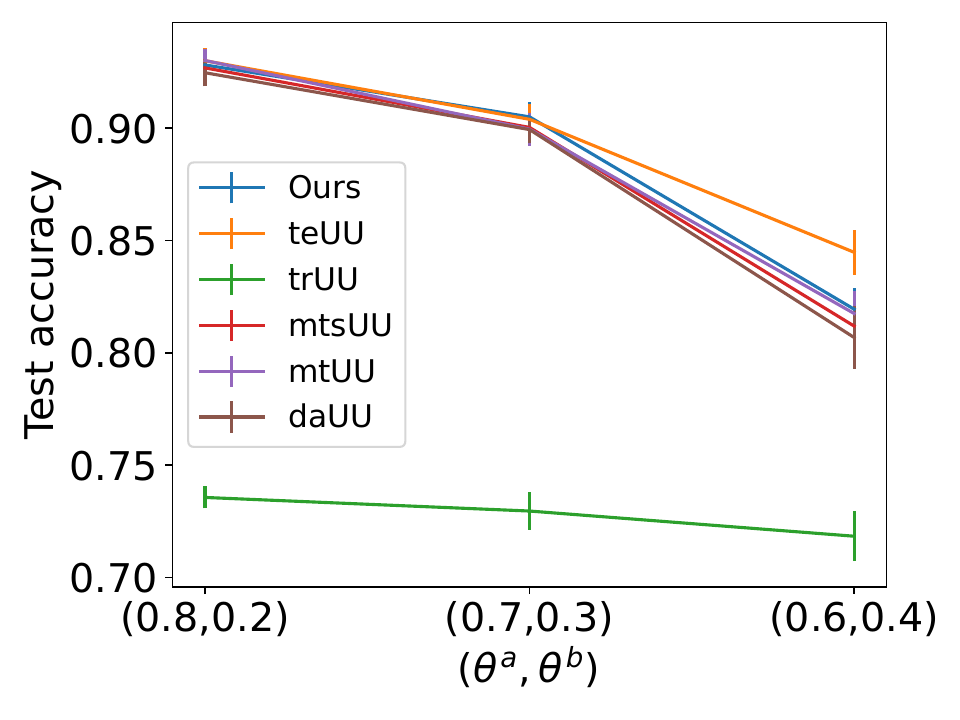}
        \subcaption{FMNIST(IO)}
    \end{minipage}
    \hfill
    \begin{minipage}{0.3\linewidth}
        \centering
        \includegraphics[width=3.4cm]{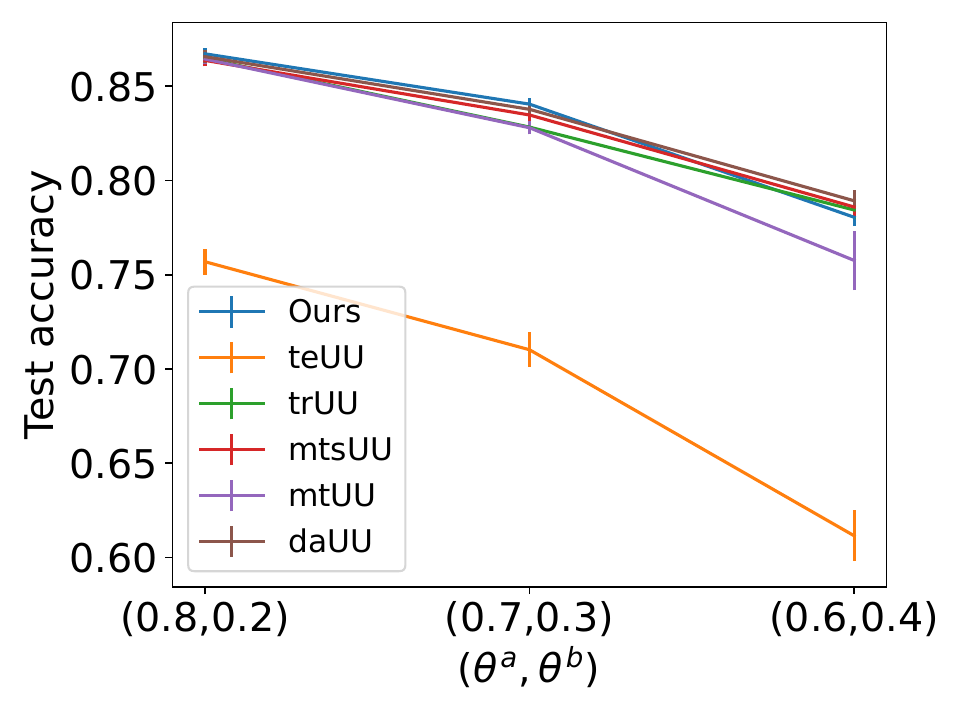}
        \subcaption{CIFAR10(S)}
    \end{minipage}
    \begin{minipage}{0.3\linewidth}
        \centering
        \includegraphics[width=3.4cm]{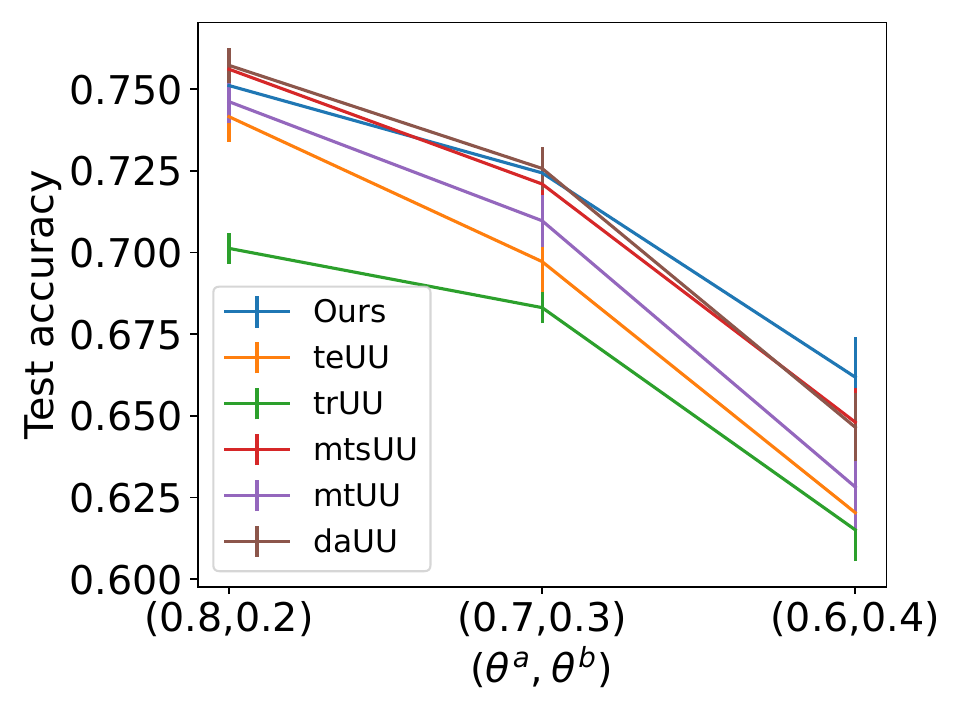}
        \subcaption{CIFAR10(IO)}
    \end{minipage}
    \begin{minipage}{0.3\linewidth}
        \centering
        \includegraphics[width=3.4cm]{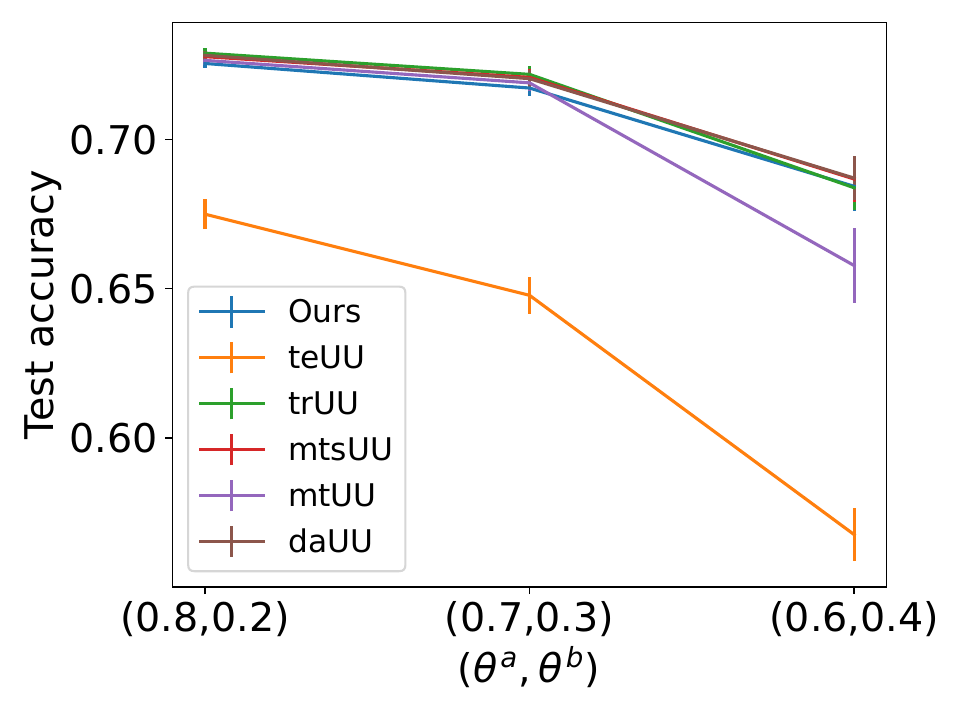}
        \subcaption{DIABETES}
    \end{minipage}
    \caption{The average test accuracies with the standard errors when changing the class-priors of UU data.}
    \label{result_theta}
\end{figure}

\subsection{Results with Different Class-priors of Given UU Datasets}
\label{dif_cps}

Figure \ref{result_theta} shows the average test accuracies with the standard errors when changing the class-priors of UU data
in the test distribution in the case where data of both the training and test distributions are noisy.
As the class-priors of UU data became closer, the performance of each method decreased.
This is because when the class-priors are close, two sets of unlabeled data become similar and thus are less informative to learn a binary classifier.
The proposed method tended to work well in each setting of the class-priors.

\begin{figure}[t]
    \centering
    \begin{minipage}{0.24\linewidth}
        \centering
        \includegraphics[width=3.4cm]{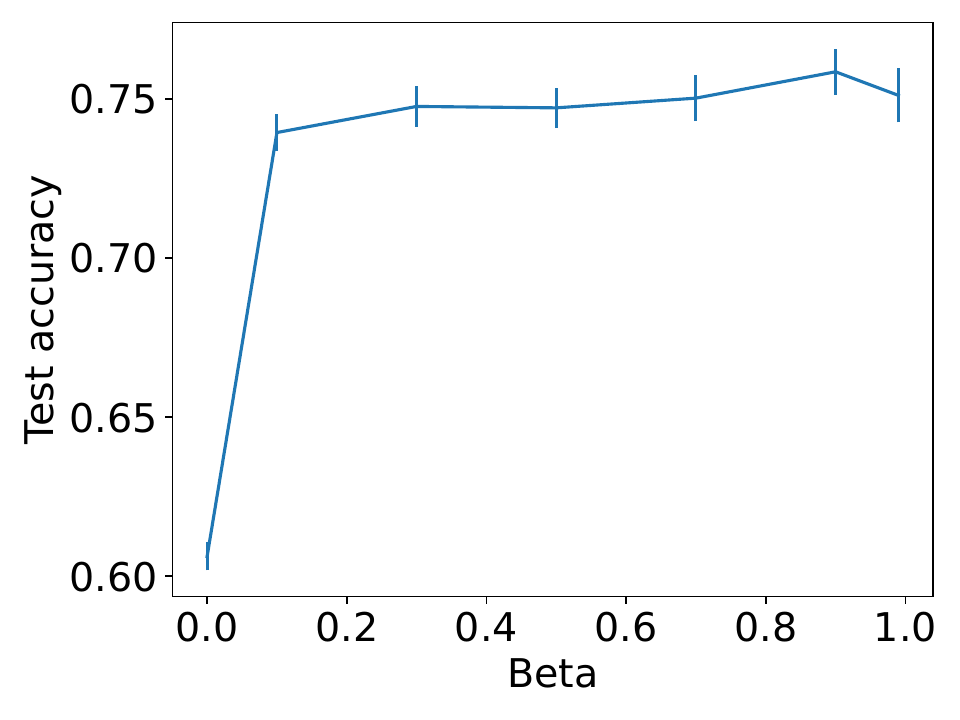}
        \subcaption{MNIST(S)}
    \end{minipage}
    \hfill
    \begin{minipage}{0.24\linewidth}
        \centering
        \includegraphics[width=3.4cm]{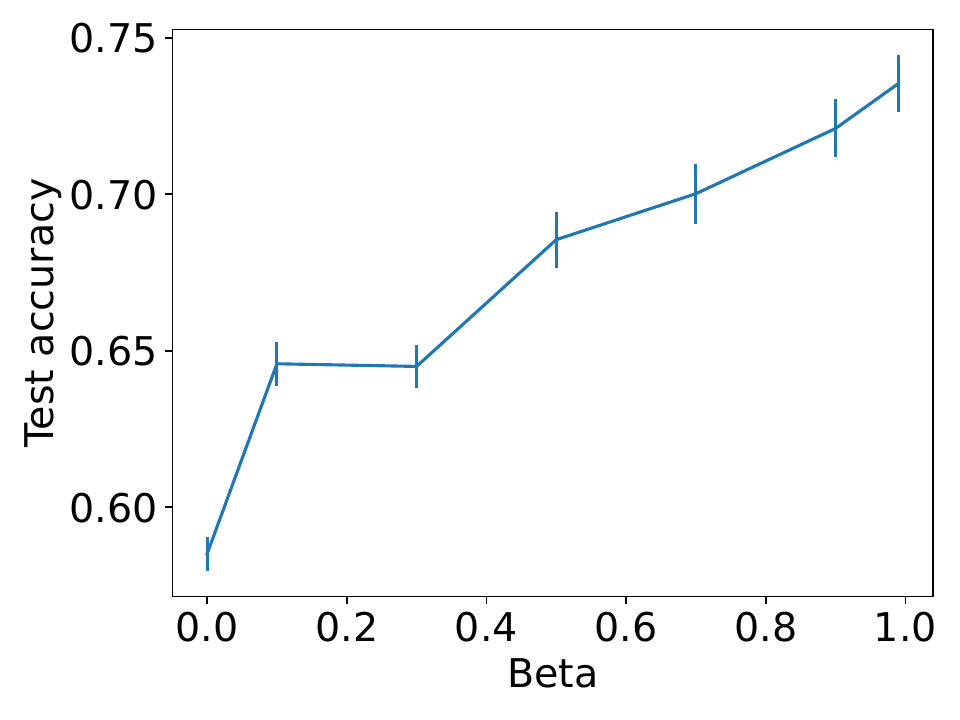}
        \subcaption{MNIST(IO)}
    \end{minipage}
    \hfill
    \begin{minipage}{0.24\linewidth}
        \centering
        \includegraphics[width=3.4cm]{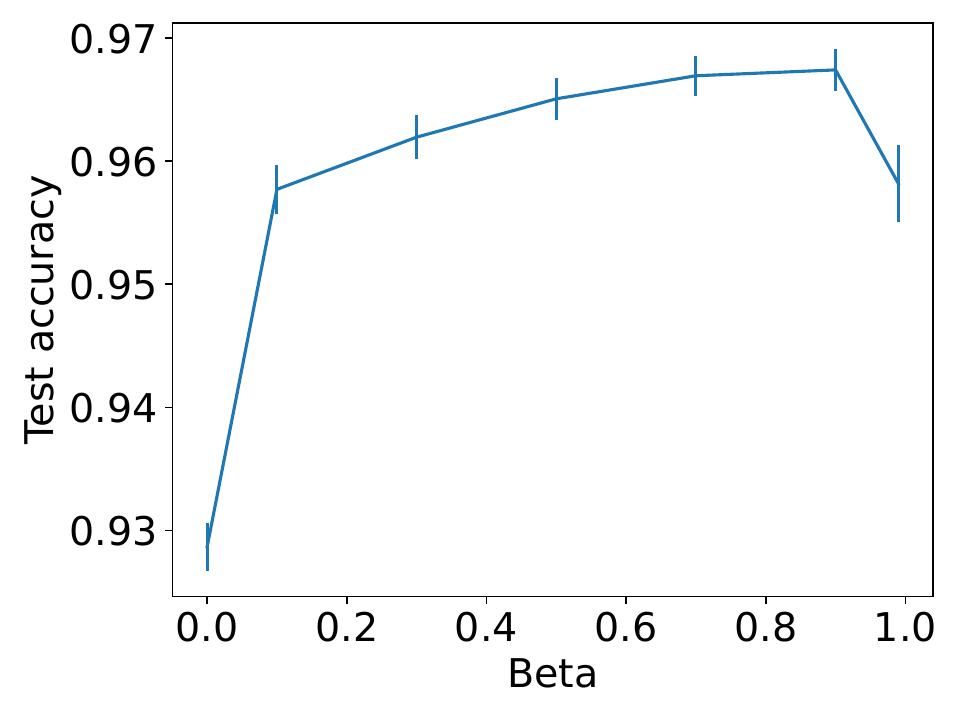}
        \subcaption{FMNIST(S)}
    \end{minipage}
    \hfill
    \begin{minipage}{0.24\linewidth}
        \centering
        \includegraphics[width=3.4cm]{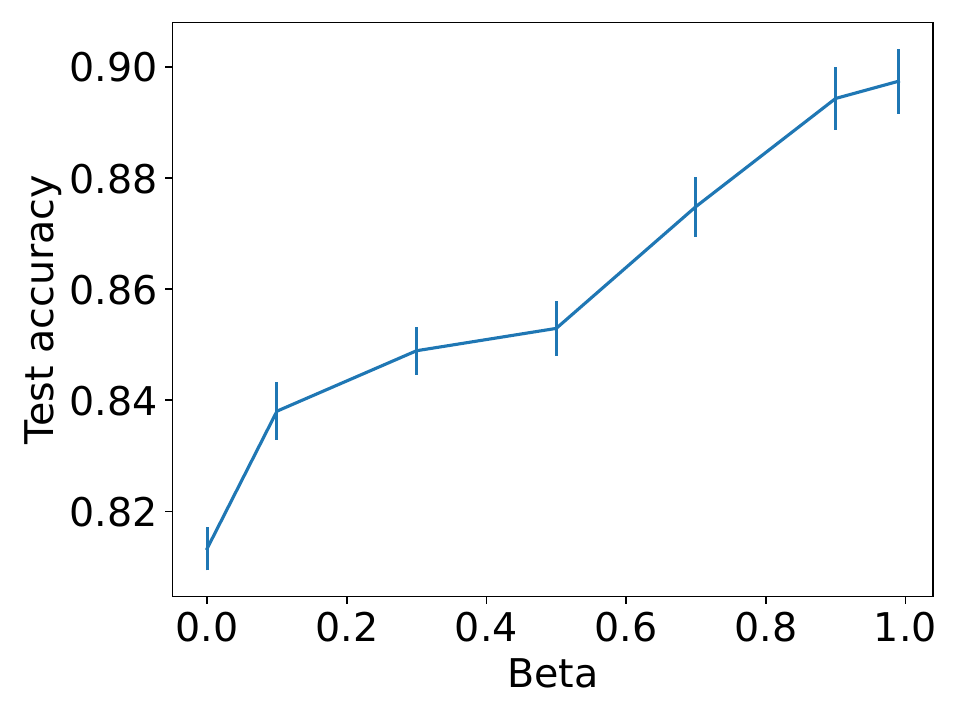}
        \subcaption{FMNIST(IO)}
    \end{minipage}
    \hfill
    \begin{minipage}{0.3\linewidth}
        \centering
        \includegraphics[width=3.4cm]{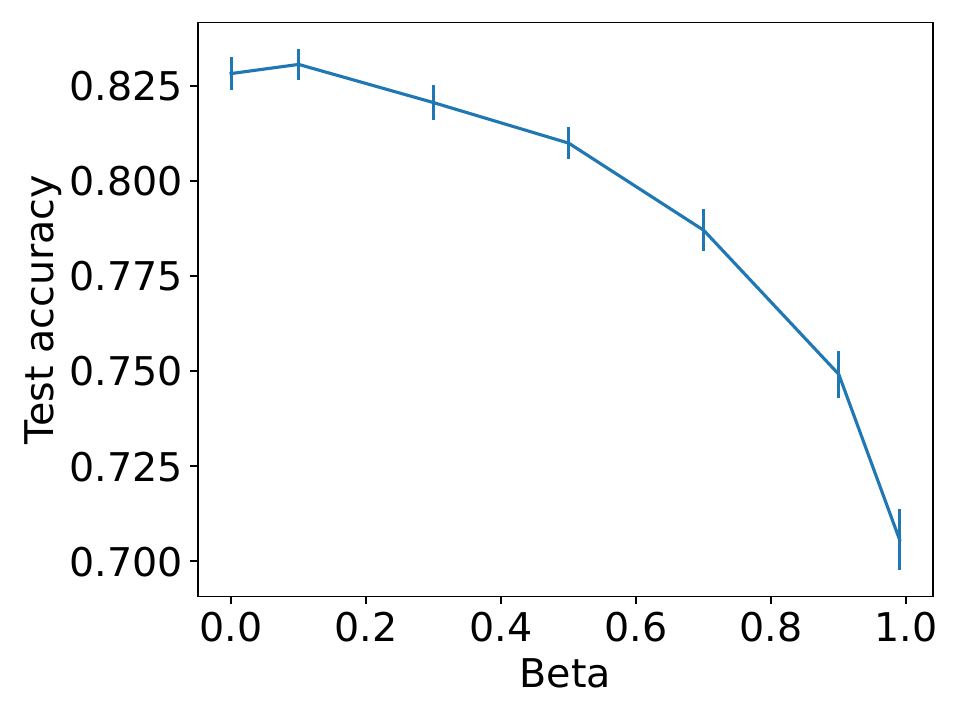}
        \subcaption{CIFAR10(S)}
    \end{minipage}
    \begin{minipage}{0.3\linewidth}
        \centering
        \includegraphics[width=3.4cm]{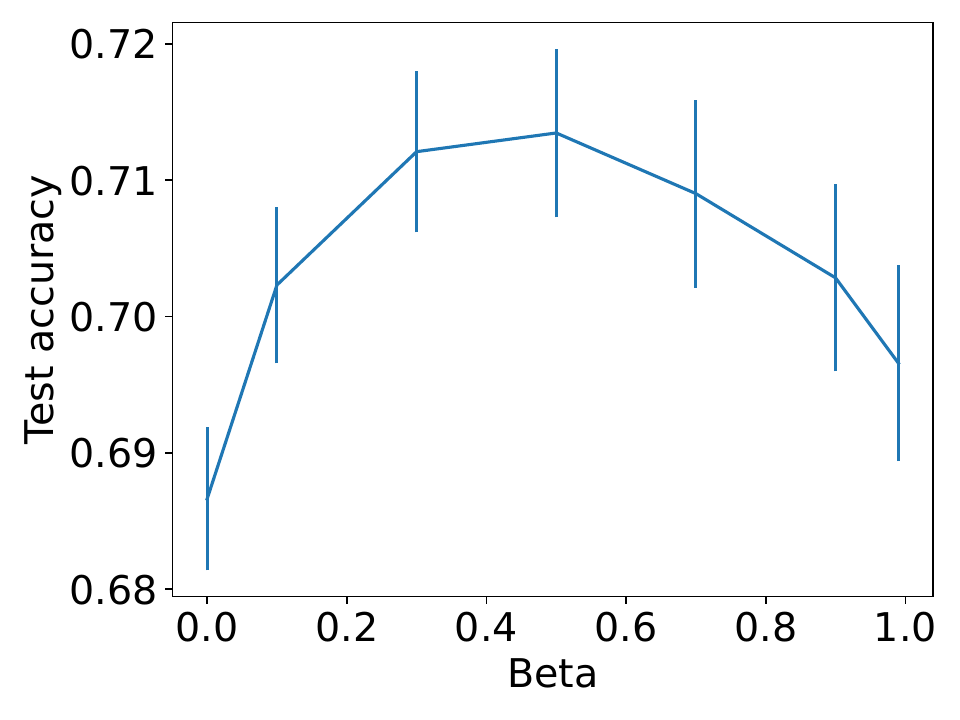}
        \subcaption{CIFAR10(IO)}
    \end{minipage}
    \begin{minipage}{0.3\linewidth}
        \centering
        \includegraphics[width=3.4cm]{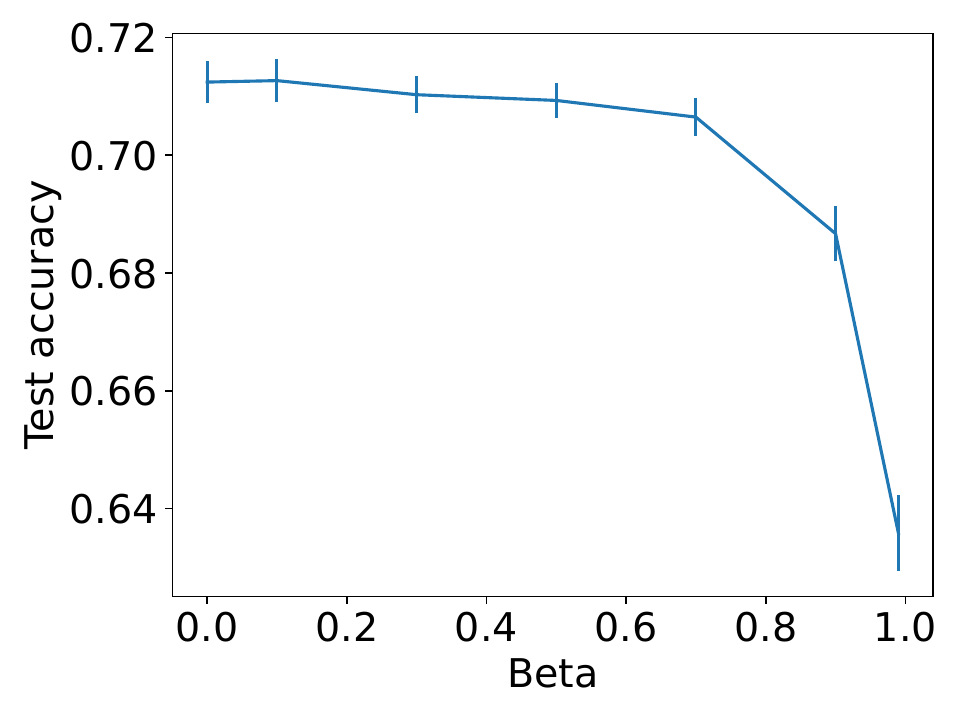}
        \subcaption{DIABETES}
    \end{minipage}
    \caption{The average test accuracies with the standard errors of our method when changing $\beta$.}
    \label{result_beta}
\end{figure}
\begin{figure}[t!]
    \centering
    \begin{minipage}{0.24\linewidth}
        \centering
        \includegraphics[width=3.4cm]{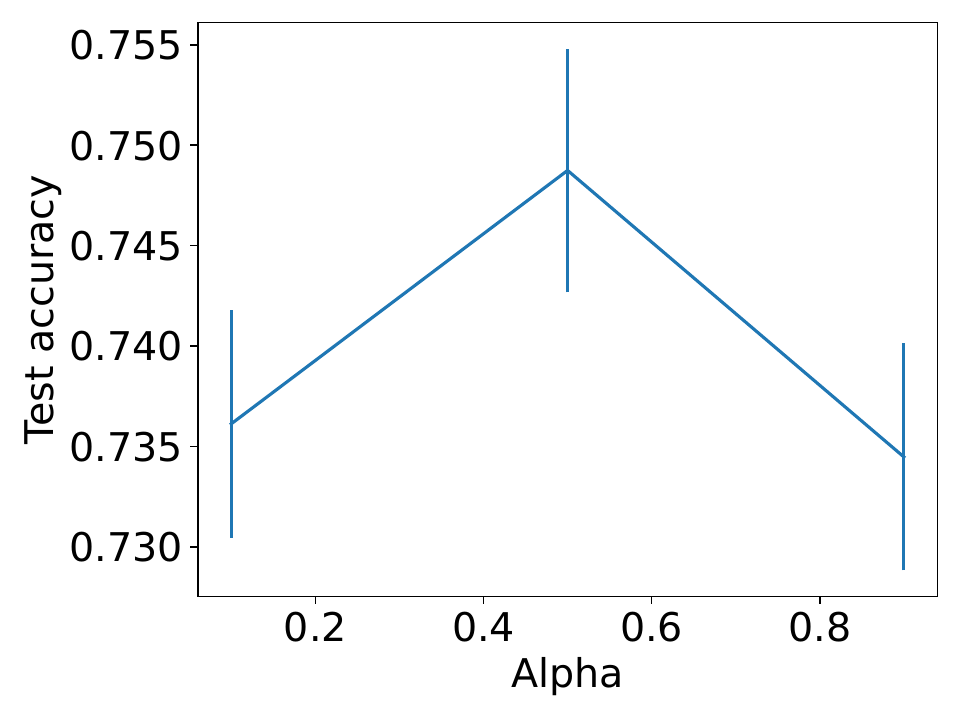}
        \subcaption{MNIST(S)}
    \end{minipage}
    \hfill
    \begin{minipage}{0.24\linewidth}
        \centering
        \includegraphics[width=3.4cm]{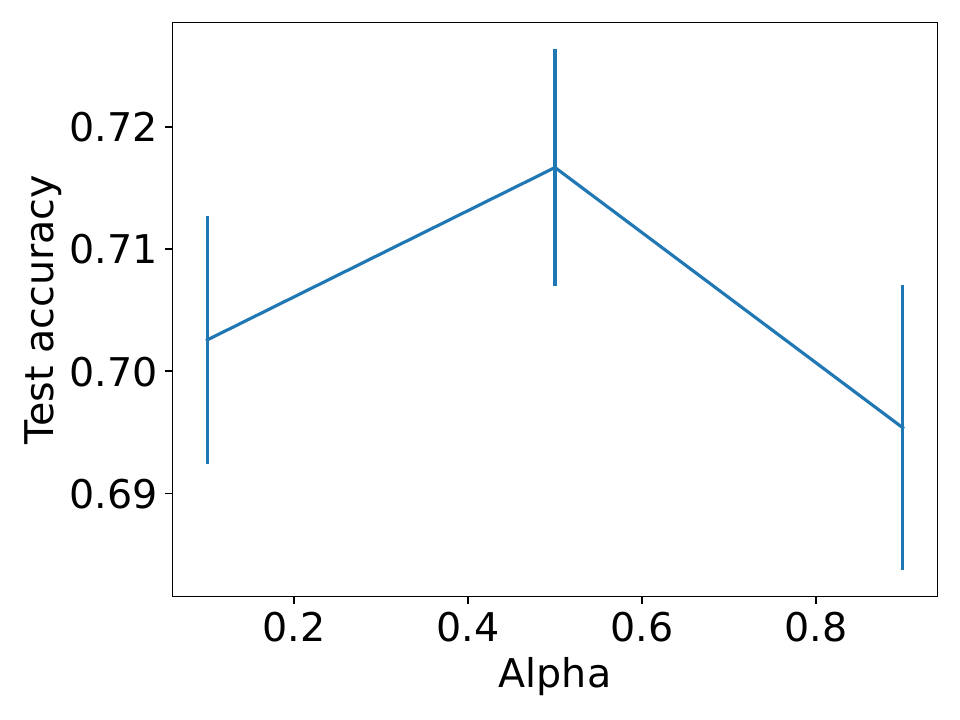}
        \subcaption{MNIST(IO)}
    \end{minipage}
    \hfill
    \begin{minipage}{0.24\linewidth}
        \centering
        \includegraphics[width=3.4cm]{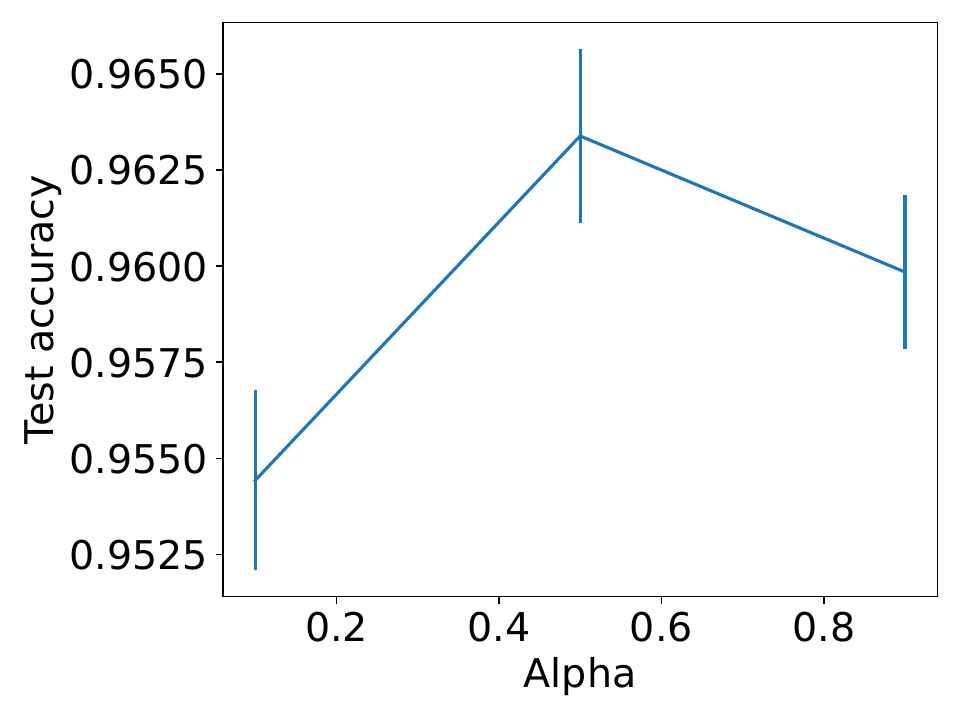}
        \subcaption{FMNIST(S)}
    \end{minipage}
    \hfill
    \begin{minipage}{0.24\linewidth}
        \centering
        \includegraphics[width=3.4cm]{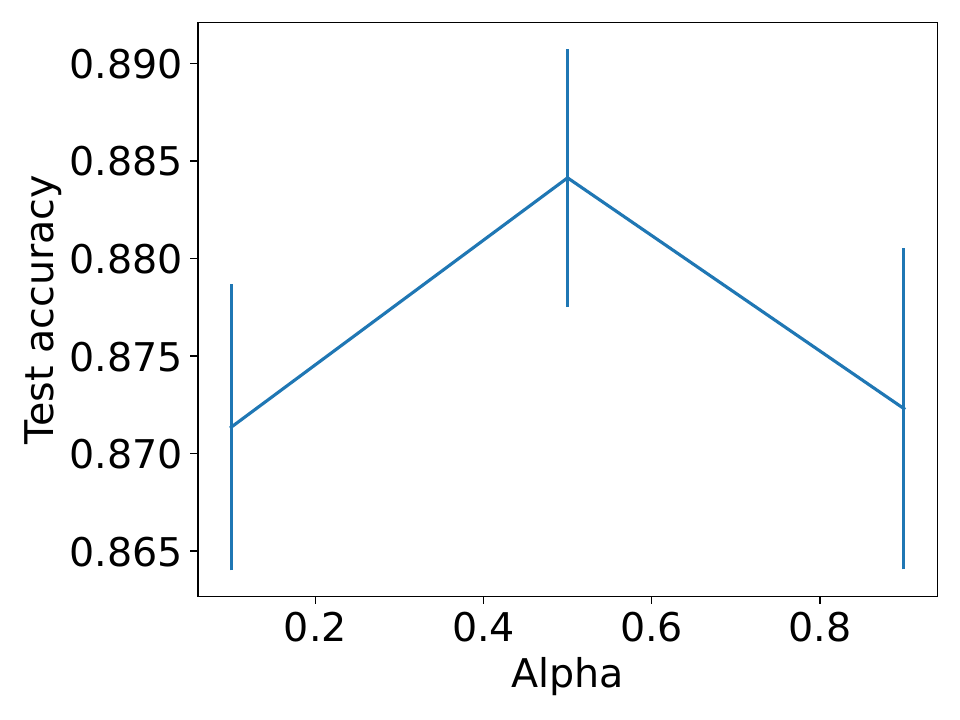}
        \subcaption{FMNIST(IO)}
    \end{minipage}
    \hfill
    \begin{minipage}{0.3\linewidth}
        \centering
        \includegraphics[width=3.4cm]{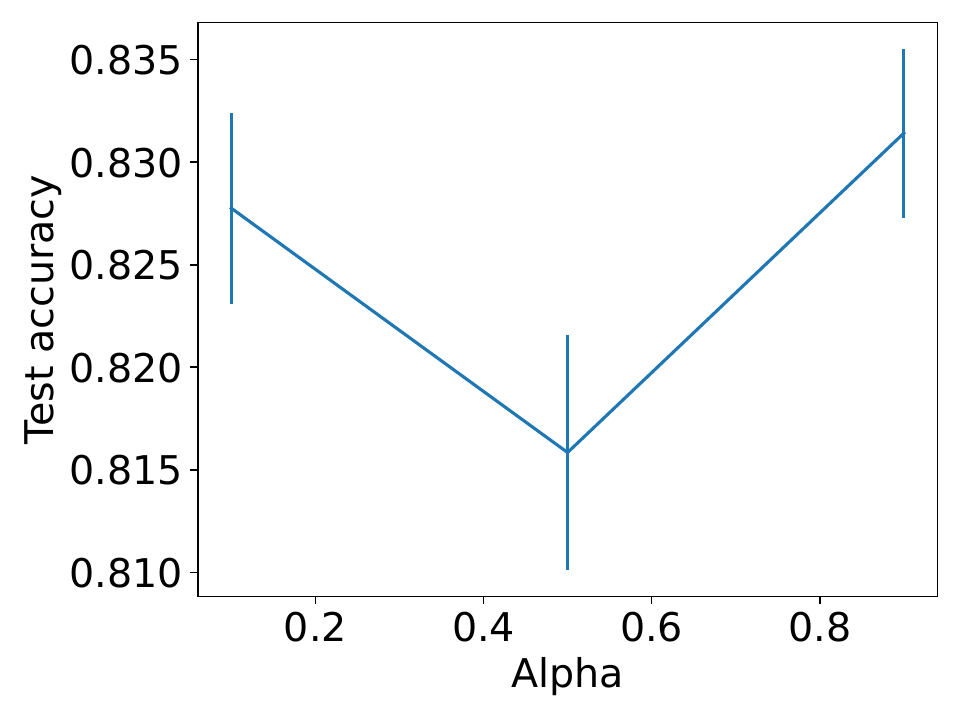}
        \subcaption{CIFAR10(S)}
    \end{minipage}
    \begin{minipage}{0.3\linewidth}
        \centering
        \includegraphics[width=3.4cm]{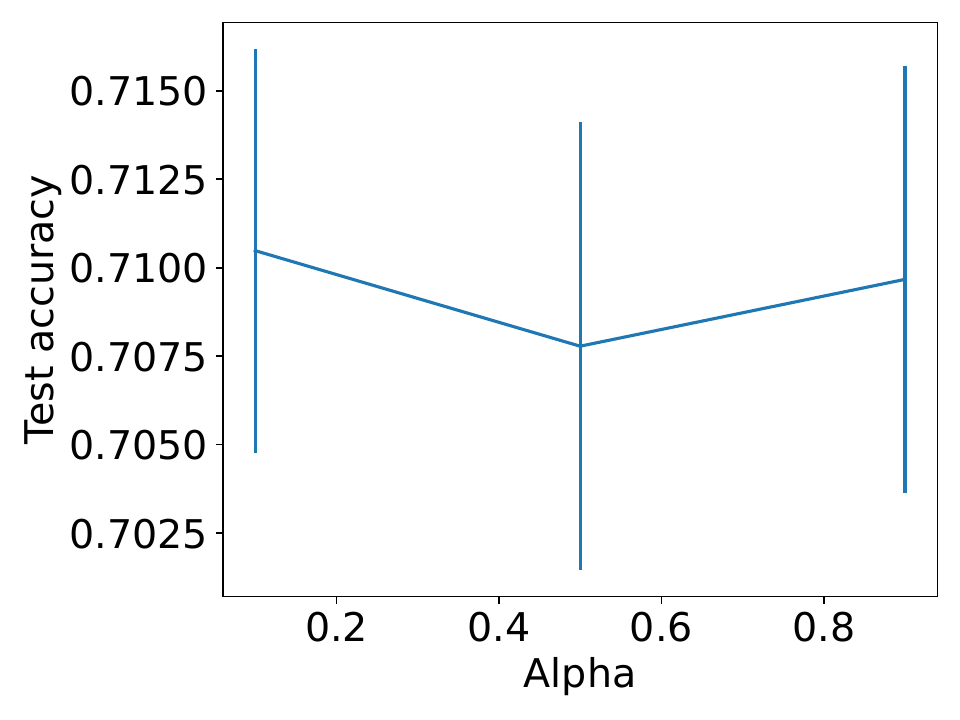}
        \subcaption{CIFAR10(IO)}
    \end{minipage}
    \begin{minipage}{0.3\linewidth}
        \centering
        \includegraphics[width=3.4cm]{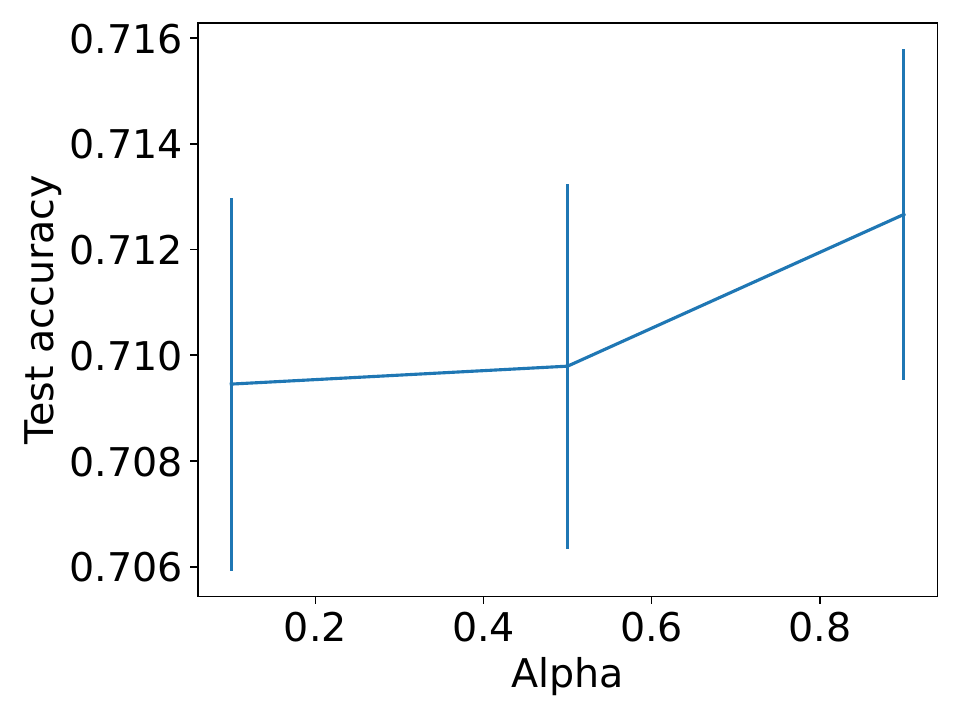}
        \subcaption{DIABETES}
    \end{minipage}
    \caption{The average test accuracies with the standard errors of our method when changing $\alpha$.}
    \label{result_alpha}
\end{figure}

\subsection{Dependency of Weighting Parameter $\beta$}
\label{depen_beta}

Figure \ref{result_beta} shows the average test accuracies with the standard errors when
changing weighting parameter $\beta$ in the case where data of both the training and test distributions are noisy.
The value of $\beta$ is used in Eq. \eqref{our_loss} for controlling the effects of two empirical risks.
Note that the proposed method with $\beta=1$ is equivalent to teUU that uses only UU data in the test distribution.
Although the trend of the results varied across datasets, the proposed method was able to select good $\beta$ by using validation UU data.

\subsection{Dependency of Relative Parameter $\alpha$}
\label{depen_alpha}

Figure \ref{result_alpha} shows the average test accuracies with the standard errors when
changing relative parameter $\alpha$ of Eq. \eqref{relative_dre} in the case where data of both the training and test distributions are noisy.
Although the trend varied across datasets, the performance tended not to change significantly even when $\alpha$ was changed.
This result shows that the proposed method is relatively robust against the value of $\alpha$.

\subsection{Comparison with Our Method that Sequentially Performs Label Estimation and Importance Weighting}

The proposed method directly performs the importance weighting from given UU datasets.
However, one possible approach is to first estimate (pseudo) PN data from the given UU data using existing UU learning methods
and then apply the importance weighting method for PN data 
\cite{fang2024generalizing}. 
We compared this approach (Ours w/ seq) with the proposed method in Table \ref{result_of_sim}. 
Note that Ours w/ seq is also considered within our proposal since importance weighting methods for UU data do not exist.
The proposed method performed comparably to or even better than Ours w/ seq.
Since Ours w/ seq sequentially performs the label estimation and the importance weighting, the error on the label estimation propagates the subsequent importance weighting, which may lead to the performance degrading in some cases.
In contrast, since the proposed method performs both simultaneously, it might be more robust.

\begin{table}[t]
\caption{Comparison with the proposed method that sequentially performs the label estimation and the importance weighting (Ours w/ seq): average test accuracies over different UU data sizes in the test distribution and class-priors.
Values in bold are not statistically different at the $5\%$ level from the best performing method in each row according to a paired t-test.
}
\label{result_of_sim}
\centering
\scalebox{1.0}{
\begin{tabular}{l|rr|rr|rr|r}
\hline
Data & MNIST & & FMNIST & & CIFAR10 & & DIABETES  \\
Shift & S & IO & S & IO & S & IO & \\
\hline
Ours & \bf{0.7482} & \bf{0.7211} & \bf{0.9620} & \bf{0.8841} & \bf{0.8294} & \bf{0.7125} & \bf{0.7090} \\
Ours w/ seq & \bf{0.7515} & \bf{0.7186} & \bf{0.9593} & \bf{0.8906} & 0.8170 & 0.6959 & \bf{0.7123} \\
\hline
\end{tabular}
}
\end{table}

\subsection{Results with Noisy Class-priors}
\label{depen_noise}

Hitherto, we have assumed that true class-priors of given UU datasets $\theta^{{\rm a}}_{{\rm tr}}$, $\theta^{{\rm b}}_{{\rm tr}}$, $\theta^{{\rm a}}_{{\rm te}}$, and $\theta^{{\rm b}}_{{\rm te}}$ are known to train classifiers.
However, they might be unavailable in practice.
Therefore, we investigated the performance of the proposed method with noisy class-priors. 

We first considered independent noise in the class-priors 
following the previous work
\cite{lu2021binary}.
Specifically, we first set true class-priors $(\theta^{\rm a}, \theta^{\rm b}):= (\theta^{{\rm a}}_{{\rm tr}}, \theta^{{\rm b}}_{{\rm tr}})=(\theta^{{\rm a}}_{{\rm te}}, \theta^{{\rm b}}_{{\rm te}})$ for the data generation to $(0.7, 0.3)$.
Then, we replaced each $\theta \in \{ \theta^{{\rm a}}_{{\rm tr}}, \theta^{{\rm b}}_{{\rm tr}}, \theta^{{\rm a}}_{{\rm te}}, \theta^{{\rm b}}_{{\rm te}} \}$ by $\theta'=\theta + \gamma \sigma$ where 
$\gamma$ uniform randomly takes a value in $\{-1,+1\}$ and $\sigma \in \{ 0.05, 0.1, 0.15, 0.195 \}$.
We used each noisy $\theta'$ as the class-prior for training classifiers.
Table \ref{result_noisy_pr} shows the average test accuracies when changing values of $\sigma$.
Here, $\sigma=0$ means that there is no noise.
As the value of $\sigma$ (magnitude of noise) increased, the performance of the proposed method tended to decrease.
This result is reasonable since information on class-priors is only supervision to train a classifier.

We next considered a simple correlated/systematic noise setting, in which all class-priors of given UU datasets, $\theta^{{\rm a}}_{{\rm tr}}$, $\theta^{{\rm b}}_{{\rm tr}}$, $\theta^{{\rm a}}_{{\rm te}}$, and $\theta^{{\rm b}}_{{\rm te}}$, are jointly biased in the same direction rather than perturbed independently.
Specifically, we replaced each $\theta$ by $\theta' = \theta + \sigma$.
We chose this setting because, in practice, estimates or prior knowledge may tend to be consistently overestimated or underestimated, rather than perturbed independently.
Table \ref{tab:correlated_noise} shows the average test accuracies when changing values of $\sigma$.
As the noise magnitude $|\sigma|$ increased, the performance of the proposed method also tended to decrease.
However, compared with the independent-noise result in Table \ref{result_noisy_pr}, 
the degradation was overall more moderate (e.g., on CIFAR10 (S) with $\sigma=0.195$, the test accuracy is $0.8229$ in Table~\ref{tab:correlated_noise}, compared with $0.7611$ in Table~\ref{result_noisy_pr}). A possible reason is that this noise shifts all class-priors in the same direction and thus tends to preserve their relative relationships, leading to a milder degradation.

\begin{table}[t]
\caption{Results of the proposed method with independent noise in the class-priors: average test accuracies over different UU data sizes in the test distribution. $\sigma$ represents the magnitude of noise.
}
\label{result_noisy_pr}
\centering
\scalebox{1.0}{
\begin{tabular}{llrrrrrrr}
\hline
Data & Shift & $\sigma=0$ & $\sigma=0.05$ & $\sigma=0.1$ & $\sigma=0.15$ & $\sigma=0.195$  \\
\hline
MNIST & S & 0.7569 & 0.7495 & 0.7409 & 0.7245 & 0.6970   \\
 & IO & 0.7438 & 0.6968 & 0.6853 & 0.6735 & 0.6449  \\
\hline
FMNIST & S & 0.9653 & 0.9477 & 0.9391 & 0.9210 & 0.9072  \\
 & IO & 0.9050 & 0.8891 & 0.8737 & 0.8517 & 0.8371 &  \\
\hline
CIFAR10 & S & 0.8404 & 0.8345 & 0.8080 & 0.7915 & 0.7611   \\
 & IO & 0.7244 & 0.7181 & 0.7058 & 0.7095 & 0.6889 & \bf  \\
\hline
DIABETES & & 0.7172 & 0.7205 & 0.7233 & 0.7214 & 0.6916  \\
\hline
\end{tabular}
}
\end{table}

\begin{table*}[t!]
\centering
\caption{Results of the proposed method with correlated noise in the class-priors: average test accuracies over different UU data sizes in the test distribution. $|\sigma|$ represents the magnitude of noise.}
\label{tab:correlated_noise}
\scalebox{0.7}{
\begin{tabular}{llccccccccc}
\toprule
Data & Shift & $\sigma=-0.195$ & $\sigma=-0.15$ & $\sigma=-0.1$ & $\sigma=-0.05$ & $\sigma=0$ & $\sigma=0.05$ & $\sigma=0.1$ & $\sigma=0.15$ & $\sigma=0.195$ \\
\midrule
MNIST    & S  & 0.7198 & 0.7266 & 0.7443 & 0.7549 & 0.7569 & 0.7576 & 0.7469 & 0.7473 & 0.7436 \\
    & IO & 0.7003 & 0.7223 & 0.7374 & 0.7427 & 0.7438 & 0.7237 & 0.7167 & 0.6861 & 0.6750 \\
\hline
FMNIST   & S  & 0.8966 & 0.9106 & 0.9238 & 0.9481 & 0.9653 & 0.9562 & 0.9332 & 0.9221 & 0.9058 \\
   & IO & 0.8697 & 0.8701 & 0.8747 & 0.8872 & 0.9050 & 0.8918 & 0.8696 & 0.8534 & 0.8254 \\
\hline
CIFAR10  & S  & 0.8195 & 0.8246 & 0.8315 & 0.8392 & 0.8404 & 0.8399 & 0.8351 & 0.8305 & 0.8229 \\
  & IO & 0.7045 & 0.7053 & 0.7142 & 0.7167 & 0.7244 & 0.7220 & 0.7176 & 0.7105 & 0.7116 \\
\hline
DIABETES &  & 0.7206 & 0.7231 & 0.7196 & 0.7171 & 0.7172 & 0.7173 & 0.7192 & 0.7214 & 0.7194 \\
\bottomrule
\end{tabular}
}
\end{table*}

\subsection{Results on Larger-scale Image Datasets}

Here, we evaluated the proposed method on larger versions of the image datasets used in the main paper (i.e., MNIST, FMNIST, and CIFAR10).
For each image dataset, we set the training UU data size to $(N_{{\rm tr}}^{{\rm a}}, N_{{\rm tr}}^{{\rm b}})=(6,000, 6,000)$ and the test UU data size to $(N_{{\rm te}}^{{\rm a}}, N_{{\rm te}}^{{\rm b}})=(100, 100)$.
Table \ref{result_large_image_ad} shows the results.
The proposed method performed the best or comparably to it in almost all cases (5 out of 6 cases).
The results show the effectiveness of the proposed method on the image datasets with increased data sizes.

\begin{table}[t]
\caption{Results on larger-scale image data: average test accuracies over different class-priors $(\theta^{{\rm a}},\theta^{{\rm b}})$ within $\{(0.8, 0.2), (0.7, 0.3), (0.6, 0.4)\}$.
Values in bold are not statistically different at the $5\%$ level from the best performing method in each row according to a paired t-test. 
}
\label{result_large_image_ad}
\centering
\begin{tabular}{llrrrrrr}
\hline
Data & Shift & Ours & teUU & trUU & mtsUU & mtUU & daUU  \\
\hline
MNIST & S & \bf{0.7517} & \bf{0.7511} & 0.5961 & 0.7530 & \bf{0.7380} & \bf{0.7625} \\
 & IO & \bf{0.7284} & \bf{0.7394} & 0.5429 & 0.6994 & 0.7086 & 0.6982 \\
FMNIST & S & \bf{0.9700} & 0.9595 & 0.9256 & 0.9589 & 0.9511 & 0.9585 \\
 & IO & 0.8842 & \bf{0.9045} & 0.7353 & 0.8808 & \bf{0.8934} & 0.8841 \\
CIFAR10 & S & \bf{0.8481} & 0.7054 & \bf{0.8542} & \bf{0.8511} & \bf{0.8500} & \bf{0.8521} \\
 & IO & \bf{0.7140} & 0.6922 & 0.6938 & \bf{0.7236} & 0.6956 & \bf{0.7291} \\
\hline
\end{tabular}
\end{table}

\begin{table}[t]
\caption{Results on larger-scale tabular data with distribution shift: average test accuracies over different class-priors $(\theta^{{\rm a}},\theta^{{\rm b}})$ within $\{(0.8, 0.2), (0.7, 0.3), (0.6, 0.4)\}$.
Values in bold are not statistically different at the $5\%$ level from the best performing method in each row according to a paired t-test.
}
\label{result_large}
\centering
\scalebox{1.0}{
\begin{tabular}{lrrrrrr}
\hline
Data & Ours & teUU & trUU & mtsUU & mtUU & daUU  \\
\hline
DIABETES (large) & \bf{0.7240} & 0.6458 & \bf{0.7261} & \bf{0.7261} & 0.7199 & \bf{0.7250} \\
FOODSTAMP & \bf{0.7159} & 0.6314 & \bf{0.7180} & 0.7136 & 0.6909 & \bf{0.7170} \\
\hline
\end{tabular}
}
\end{table}

\begin{table*}[t]
\caption{Training time [s] of the proposed method on MNIST(S).}
\label{result_ttime}
\centering
\scalebox{1.0}{
\begin{tabular}{rrrrr}
\hline
\multicolumn{1}{c}{Ours} & \multicolumn{1}{c}{mtsUU} & \multicolumn{1}{c}{mtUU} & \multicolumn{1}{c}{daUU} \\
\hline
118.522 & 108.0755 & 107.771 & 143.969 \\
\hline
\end{tabular}
}
\end{table*}

\begin{table*}[t]
\caption{Average F1 scores over different UU test data sizes $(N_{{\rm te}}^{{\rm a}}, N_{{\rm te}}^{{\rm b}})$ within $ \{ (50, 50), (100, 100), (150, 150) \} $ and class-priors $(\theta^{{\rm a}},\theta^{{\rm b}})$ within $\{(0.8, 0.2), (0.7, 0.3), (0.6, 0.4)\}$.
In the Shift column, `S' and `IO' represent the support shift and input-output relation shift, respectively.
Values in bold are not statistically different at the $5\%$ level from the best performing method in each row according to a paired t-test.
}
\label{f1_result}
\centering
\scalebox{1.0}{
\begin{tabular}{llrrrrrrrr}
\hline
Data & Shift & \multicolumn{1}{c}{Ours} & \multicolumn{1}{c}{teUU} & \multicolumn{1}{c}{trUU} & \multicolumn{1}{c}{mtsUU} &  \multicolumn{1}{c}{mtUU} &
\multicolumn{1}{c}{daUU} \\
\hline
MNIST    & S     & \textbf{0.7328} & \textbf{0.7325} & 0.5038 & \textbf{0.7253} & \textbf{0.7317} & \textbf{0.7295} \\
         & IO    & \textbf{0.7170} & \textbf{0.7281} & 0.4747 & 0.6752 & 0.6858 & 0.6676 \\
\hline
FMNIST   & S     & \textbf{0.9610} & 0.9527 & 0.9107 & 0.9537 & 0.9492 & 0.9537 \\
         & IO    & \textbf{0.8793} & \textbf{0.8889} & 0.7327 & 0.8745 & 0.8763 & 0.8711 \\
\hline
CIFAR10  & S     & \textbf{0.8288} & 0.6912 & 0.8220 & \textbf{0.8244} & 0.8076 & \textbf{0.8283} \\
         & IO    & \textbf{0.7174} & 0.6873 & 0.6501 & \textbf{0.7058} & 0.6964 & \textbf{0.7074} \\
\hline
DIABETES & –     & \textbf{0.7186} & 0.6368 & \textbf{0.7203} & \textbf{0.7210} & \textbf{0.7019} & \textbf{0.7219} \\
\hline
\end{tabular}
}
\end{table*}

\subsection{Results on Larger-scale Tabular Datasets with Real Distribution Shift}

We evaluated the proposed method with larger-scale tabular datasets with real distribution shift.
In this experiments, we newly used a FOODSTAMP, a real-world tabular dataset used for recent distribution adaptation studies
\cite{gardner2024benchmarking,kumagaiimportance}. 
In this dataset, the task is to predict whether an individual is receiving food stamps. The distribution shift occurs due to the difference of the geographic regions in which individuals live.
For FOODSTAMP and DIABETES datasets, we set the training UU data size $(N_{{\rm tr}}^{{\rm a}}, N_{{\rm tr}}^{{\rm b}})=(6,000, 6,000)$ and the test UU data size $(N_{{\rm te}}^{{\rm a}}, N_{{\rm te}}^{{\rm b}})=(150, 150)$ to evaluate the proposed method with a larger-scale regime.
Table \ref{result_large} shows the results.
The proposed method remains competitive in such larger-scale cases.

\subsection{Computation Costs}

We evaluated the training time of the proposed method on MNIST(S) with $(N_{{\rm te}}^{{\rm a}}, N_{{\rm te}}^{{\rm b}})=(150, 150)$. 
We used a Linux server with a 2.20Hz CPU.
For comparison, we also evaluated the methods that use both UU data in the training and the test distributions as in the proposed method.
Table \ref{result_ttime} shows the results. Since the proposed method estimated both importance weights and classifiers, it had slightly longer training time than mtsUU and mtUU. 
However, the differences were not significant. 
daUU had longer training time than the proposed method due to the calculation of the MMD loss to mitigate the feature discrepancy.
This result indicates that the proposed method is practical in terms of computation costs.

\subsection{F1 Scores}
\label{f1_score}

Although we used test accuracies as the primary evaluation metric in the main paper, alternative metrics, such as F1 scores, can also be informative.
Table \ref{f1_result} shows the average F1 scores for each dataset under the setting where data of both the training and test distributions are noisy.
The proposed method also performed well in terms of the F1 scores.

\subsection{Full Results with Standard Deviations}
\label{full_dev}

Tables \ref{result_all_std} and \ref{result_all_diftheta2_std} show the average test accuracies with their standard deviations of each method.
 
\begin{table}[t]
\caption{Average test accuracies with standard deviations over different UU test data sizes $(N_{{\rm te}}^{{\rm a}}, N_{{\rm te}}^{{\rm b}})$ within $ \{ (50, 50), (100, 100), (150, 150) \} $ and class-priors $(\theta^{{\rm a}},\theta^{{\rm b}})$ within $\{(0.8, 0.2), (0.7, 0.3), (0.6, 0.4)\}$.
In the Shift column, `S' and `IO' represent the support shift and input-output relation shift, respectively.
Values in bold are not statistically different at the $5\%$ level from the best performing method in each row according to a paired t-test.
}
\label{result_all_std}
\centering
\scalebox{0.75}{
\begin{tabular}{llrrrrrrrr}
\hline
Data & Shift & \multicolumn{1}{c}{Ours} & \multicolumn{1}{c}{teUU} & \multicolumn{1}{c}{trUU} & \multicolumn{1}{c}{mtsUU} &  \multicolumn{1}{c}{mtUU} &
\multicolumn{1}{c}{daUU} \\
\hline
MNIST & S & \bf{0.7482(0.059)} & \bf{0.7419(0.083)} & 0.5879(0.035) & 0.7388(0.062) & \bf{0.7355(0.094)} & 0.7423(0.060) \\
 & IO & 0.7211(0.089) & \bf{0.7334(0.083)} & 0.5412(0.025) & 0.6922(0.110) & 0.6996(0.116) & 0.6929(0.113)  \\
\hline
FMNIST & S & \bf{0.9620(0.019)} & 0.9531(0.040) & 0.9161(0.015) & 0.9551(0.022) & 0.9509(0.029) & 0.9550(0.022) \\
 & IO & \bf{0.8841(0.063)} & \bf{0.8929(0.055)} & 0.7278(0.047) & 0.8796(0.069) & 0.8823(0.064) & 0.8769(0.072) \\
\hline
CIFAR10 & S & \bf{0.8294(0.042)} & 0.6929(0.082) & 0.8258(0.039) & 0.8281(0.038) & 0.8166(0.068) & \bf{0.8308(0.039)} \\
 & IO & \bf{0.7125(0.061)} & 0.6864(0.077) & 0.6665(0.052) & \bf{0.7084(0.060)} & 0.6947(0.072) & \bf{0.7099(0.063)} \\
\hline
DIABETES & & \bf{0.7090(0.033)} & 0.6300(0.059) & \bf{0.7115(0.033)} & \bf{0.7117(0.032)} & 0.7011(0.051) & \bf{0.7118(0.031)} \\
\hline
\end{tabular}
}
\end{table}
\begin{table}[t!]
\caption{Results with a few PN data in the test distribution: Average test accuracies with standard deviations over different UU test data sizes $(N_{{\rm te}}^{{\rm a}}, N_{{\rm te}}^{{\rm b}})$ within $ \{ (10,10), (50, 50), (100, 100) \} $ and training class-priors $(\theta_{{\rm tr}}^{{\rm a}},\theta_{{\rm tr}}^{{\rm b}})$ within $\{(0.8, 0.2), (0.7, 0.3), (0.6, 0.4)\}$. Class-priors of UU test data $(\theta_{{\rm te}}^{{\rm a}},\theta_{{\rm te}}^{{\rm b}})$ were set to $(1,0)$.
}
\label{result_all_diftheta2_std}
\centering
\scalebox{0.75}{
\begin{tabular}{llrrrrrrrr}
\hline
Data & Shift & \multicolumn{1}{c}{Ours} & \multicolumn{1}{c}{teUU} & \multicolumn{1}{c}{trUU} & \multicolumn{1}{c}{mtsUU} &  \multicolumn{1}{c}{mtUU} &
\multicolumn{1}{c}{daUU} \\
\hline
MNIST & S & 0.8237(0.058) & 0.8184(0.055) & 0.6012(0.031) & 0.8194(0.057) & \bf{0.8309(0.059)} & 0.8177(0.057)  \\
 & IO & \bf{0.8075(0.057)} & \bf{0.8102(0.055)} & 0.5543(0.020) & 0.8012(0.064) & \bf{0.8127(0.063)} & 0.7998(0.074)  \\
\hline
FMNIST & S & \bf{0.9726(0.013)} & 0.9655(0.024) & 0.9181(0.015) & \bf{0.9719(0.014)} & 0.9705(0.018) & \bf{0.9717(0.014)}  \\
 & IO & \bf{0.9367(0.034)} & 0.9285(0.047) & 0.7505(0.033) & 0.9364(0.033) & 0.9299(0.048) & \bf{0.9393(0.033)}  \\
\hline
CIFAR10 & S & \bf{0.8492(0.027)} & 0.7555(0.083) & 0.8286(0.035) & 0.8422(0.028) & 0.8376(0.029) & \bf{0.8466(0.027)} \\
 & IO & \bf{0.7605(0.053)} & 0.7455(0.064) & 0.6732(0.045) & 0.7563(0.046) & 0.7492(0.055) & \bf{0.7611(0.047)}  \\
\hline
DIABETES & & \bf{0.7251(0.013)} & 0.6757(0.042) & \bf{0.7233(0.016)} & \bf{0.7252(0.011)} & 0.7209(0.015) & \bf{0.7238(0.012)}  \\
\hline
\end{tabular}
}
\end{table}


\end{document}